%% file: main.tex
\documentclass[journal]{IEEEtran}
\usepackage{amsmath,amsfonts}
\usepackage{algorithmic}
\usepackage{array}
\usepackage{textcomp}
\usepackage{stfloats}
\usepackage{url}
\usepackage{verbatim}
\usepackage{graphicx}
\def\BibTeX{{\rm B\kern-.05em{\sc i\kern-.025em b}\kern-.08em
    T\kern-.1667em\lower.7ex\hbox{E}\kern-.125emX}}
\usepackage{balance}
\usepackage[caption=false,font=normalsize,labelfont=sf,textfont=sf]{subfig}
\usepackage{cleveref}
\usepackage{siunitx}
\usepackage{multirow}

\usepackage{titlesec}
\usepackage{float}

\makeatletter
\def\ps@IEEEtitlepagestyle{%
\def\@oddfoot{\mycopyrightnotice}%
\def\@evenfoot{}%
}
\def\mycopyrightnotice{%
{\footnotesize 10.1109/TITS.2024.3386195 ~\copyright~2024 IEEE\hfill} 
\gdef\mycopyrightnotice{}
}

\begin{document}
%
\title{Evaluating Pedestrian Trajectory Prediction Methods with Respect to Autonomous Driving}

\author{Nico~Uhlemann,
        Felix~Fent,
        Markus~Lienkamp
\thanks{Nico Uhlemann, Felix Fent and Markus Lienkamp are with the Technical University of Munich, Germany; School of Engineering \& Design, Institute of Automotive Technology and Munich Institute of Robotics and Machine Intelligence (MIRMI)}
}

\markboth{© 2024 IEEE. Personal use of this material is permitted. Permission from IEEE must be obtained for all other uses.}{}%

\maketitle

\begin{abstract}
In this paper, we assess the state of the art in pedestrian trajectory prediction within the context of generating single trajectories, a critical aspect aligning with the requirements in autonomous systems. The evaluation is conducted on the widely-used ETH/UCY dataset where the Average Displacement Error (ADE) and the Final Displacement Error (FDE) are reported. Alongside this, we perform an ablation study to investigate the impact of the observed motion history on prediction performance. To evaluate the scalability of each approach when confronted with varying amounts of agents, the inference time of each model is measured. Following a quantitative analysis, the resulting predictions are compared in a qualitative manner, giving insight into the strengths and weaknesses of current approaches. The results demonstrate that although a constant velocity model (CVM) provides a good approximation of the overall dynamics in the majority of cases, additional features need to be incorporated to reflect common pedestrian behavior observed \cite{rasouli_autonomous_2018, gorrini_observation_2018}. Therefore, this study presents a data-driven analysis with the intent to guide the future development of pedestrian trajectory prediction algorithms.
\end{abstract}


\IEEEpeerreviewmaketitle

\input{chapters/01_introduction}
\input{chapters/02_related_work}

\input{chapters/03_methodology}
\input{chapters/04_results}
\input{chapters/05_discussion}
\input{chapters/06_conclusion}

\section*{Acknowledgment}
As the first author, Nico Uhlemann initiated the idea of this paper and contributed essentially to its conception, implementation, and content. Felix Fent contributed to the conception of this research, the analysis of the generated data, and the revision of the research article. Markus Lienkamp made an essential contribution to the conception of the research project and revised the paper critically for important intellectual content. He gave final approval of the version to be published and agreed with all aspects of the work. As a guarantor, he accepts responsibility for the overall integrity of the paper.
The authors would like to thank their project partner Enway GmbH, as well as the Munich Institute of Robotics and Machine Intelligence (MIRMI) for their support. The research was funded by the Central Innovation Program (ZIM) under grant No. KK5213703GR1.

\appendices
\vspace{-4.0mm}
\section{}
\label{app:eval_runtime}
\vspace{-4.0mm}
\begin{figure}[!h]
    \begin{center}\small{a) Y-Net} \\\end{center}
    \vspace{-2.0mm}
    \begin{tabular}{@{}c@{}}
        \centering
        \fontsize{8pt}{12pt}\selectfont
        \def\svgwidth{0.98\linewidth}
		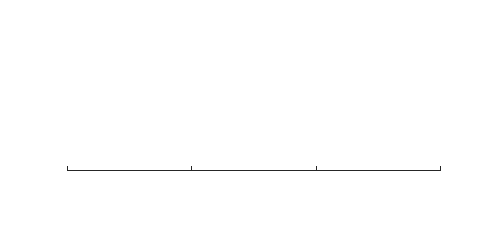
    \end{tabular}
    
    \begin{center}\small{b) Trajectron++} \\\end{center}
    \vspace{-2.0mm}
	\begin{tabular}{@{}c@{}}
        \centering
        \fontsize{8pt}{12pt}\selectfont
        \def\svgwidth{0.98\linewidth}
		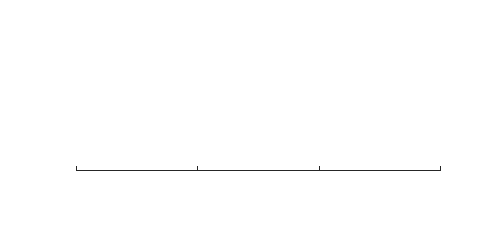
    \end{tabular}

    \vspace{-2.0mm}
    \caption{Runtime over the number of agents in a scene for the a)~Y-Net and b)~Trajectron++ model.}
    \label{fig:runtime_agents}
\end{figure}

\section{}
\label{app:eval_values}

The ADE and FDE values for the experiments conducted in \cref{sec_res} are listed in the tables below.

\begin{table*}[t]
\vspace{-2.0mm}
\centering
\caption{ADE and FDE values across all models for 20 trajectories generated (K=20) as well as only one (K=1), considering the whole motion history of N=8 timesteps}
\begin{tabular}{c|c|c|c|c|c|c|c|c|c|c|c|c}
\hline
 \multirow{2}{*}{Metric} & \multirow{2}{*}{Dataset} & \multicolumn{2}{c|}{SGAN} & \multicolumn{2}{c|}{Y-Net} & \multicolumn{2}{c|}{Trajectron++} & \multicolumn{2}{c|}{Social-Implicit} & \multicolumn{2}{c|}{AgentFormer} & CVM\\
 & & K=20 & K=1 & K=20 & K=1 & K=20 & K=1 & K=20 & K=1 & K=20 & K=1 & K=1 \\
 \hline
 \hline
\multirow{6}{*}{ADE} & ETH & 0.705 & 1.135 & 0.585 & 1.599 & 0.536 & 1.039 & 0.667 & 1.057 & 0.451 & 1.416 & 0.995\\
& Hotel & 0.478 & 0.776 & 0.225 & 0.614 & 0.162 & 0.367 & 0.202 & 0.774 & 0.142 & 0.728 & 0.323\\
& Univ & 0.558 & 0.768 & 0.539 & 1.460 & 0.282 & 0.558 & 0.310 & 0.731 & 0.254 & 0.842 & 0.524\\
& Zara1 & 0.337 & 0.638 & 0.363 & 1.125 & 0.215 & 0.468 & 0.257 & 0.567 & 0.177 & 0.744 & 0.431\\
& Zara2 & 0.307 & 0.553 & 0.284 & 0.787 & 0.161 & 0.343 & 0.222 & 0.537 & 0.140 & 0.724 & 0.326\\
\hline
& \textbf{Average} & \textbf{0.477} & \textbf{0.774} & \textbf{0.399} & \textbf{1.117} & \textbf{0.271} & \textbf{0.555} & \textbf{0.332} & \textbf{0.733} & \textbf{0.233} & \textbf{0.891} & \textbf{0.520}\\
\hline
\multirow{6}{*}{FDE} & ETH & 1.286 & 2.245 & 0.695 & 3.234 & 0.944 & 2.120 & 1.461 & 2.340 & 0.748 & 2.961 & 2.234\\
& Hotel & 1.018 & 1.683 & 0.272 & 1.207 & 0.255 & 0.670 & 0.359 & 1.475 & 0.225 & 1.610 & 0.617\\
& Univ & 1.182 & 1.681 & 0.958 & 3.087 & 0.551 & 1.237 & 0.600 & 1.495 & 0.454 & 1.846 & 1.165\\
& Zara1 & 0.685 & 1.395 & 0.570 & 2.333 & 0.412 & 1.031 & 0.509 & 1.152 & 0.304 & 1.641 & 0.960\\
& Zara2 & 0.642 & 1.205 & 0.451 & 1.651 & 0.308 & 0.754 & 0.441 & 1.117 & 0.236 & 1.647 & 0.728\\
\hline
& \textbf{Average} & \textbf{0.962} & \textbf{1.642} & \textbf{0.589} & \textbf{2.302} & \textbf{0.494} & \textbf{1.162} & \textbf{0.674} & \textbf{1.516} & \textbf{0.393} & \textbf{1.941} & \textbf{1.141}\\
 \hline
\end{tabular}
\label{table:accuracy}
\end{table*}

\begin{table*}[t]
\vspace{-2.0mm}
\centering
\caption{ADE and FDE values across all models for N=1 and N=2 timesteps while only generating one trajectory}
\begin{tabular}{c|c|c|c|c|c|c|c|c|c|c|c}
\hline
 \multirow{2}{*}{Metric} & \multirow{2}{*}{Dataset} & \multicolumn{2}{c|}{SGAN} & \multicolumn{2}{c|}{Y-Net} & \multicolumn{2}{c|}{Trajectron++} & \multicolumn{2}{c|}{Social-Implicit} & \multicolumn{2}{c}{AgentFormer} \\
 & & N=1 & N=2 & N=1 & N=2 & N=1 & N=2 & N=1 & N=2 & N=1 & N=2 \\
 \hline
 \hline
\multirow{6}{*}{ADE} & ETH & 1.121 & 1.148 & 5.805 & 2.182 & 0.980 & 1.000 & 1.389 & 1.021 & 2.510 & 1.371\\
& Hotel & 0.722 & 0.726 & 2.370 & 0.835 & 0.381 & 0.380 & 0.993 & 0.761 & 1.498 & 1.071\\
& Univ & 0.774 & 0.772 & 2.744 & 1.629 & 0.601 & 0.583 & 0.985 & 0.731 & 1.475 & 0.924\\
& Zara1 & 0.641 & 0.648 & 5.027 & 1.728 & 0.503 & 0.470 & 1.352 & 0.654 & 2.567 & 1.417\\
& Zara2 & 0.526 & 0.524 & 2.841 & 1.104 & 0.361 & 0.362 & 0.832 & 0.560 & 1.656 & 1.092\\
\hline
& \textbf{Average} & \textbf{0.757} & \textbf{0.764} & \textbf{3.757} & \textbf{1.496} & \textbf{0.566} & \textbf{0.559} & \textbf{1.110} & \textbf{0.745} & \textbf{1.941} & \textbf{1.175}\\
\hline
\multirow{6}{*}{FDE} & ETH & 2.167 & 2.233 & 9.998 & 4.370 & 2.012 & 2.088 & 2.415 & 2.179 & 4.509 & 2.774\\
& Hotel & 1.573 & 1.586 & 4.352 & 1.803 & 0.755 & 0.755 & 1.878 & 1.453 & 2.955 & 2.246\\
& Univ & 1.696 & 1.693 & 5.005 & 3.367 & 1.331 & 1.294 & 1.842 & 1.462 & 2.788 & 1.943\\
& Zara1 & 1.381 & 1.400 & 9.269 & 3.726 & 1.101 & 1.040 & 2.588 & 1.406 & 4.810 & 2.886\\
& Zara2 & 1.136 & 1.130 & 5.261 & 2.442 & 0.810 & 0.806 & 1.646 & 1.177 & 3.245 & 2.329\\
\hline
& \textbf{Average} & \textbf{1.591} & \textbf{1.608} & \textbf{6.777} & \textbf{3.142} & \textbf{1.202} & \textbf{1.197} & \textbf{2.074} & \textbf{1.535} & \textbf{3.661} & \textbf{2.436}\\
 \hline
\end{tabular}
\label{table:accuracy}
\end{table*}

\bibliographystyle{IEEEtran}
\bibliography{bibtex/bib/my_bib}

\begin{IEEEbiography}[{\includegraphics[width=1in,height=1.25in,clip,keepaspectratio]{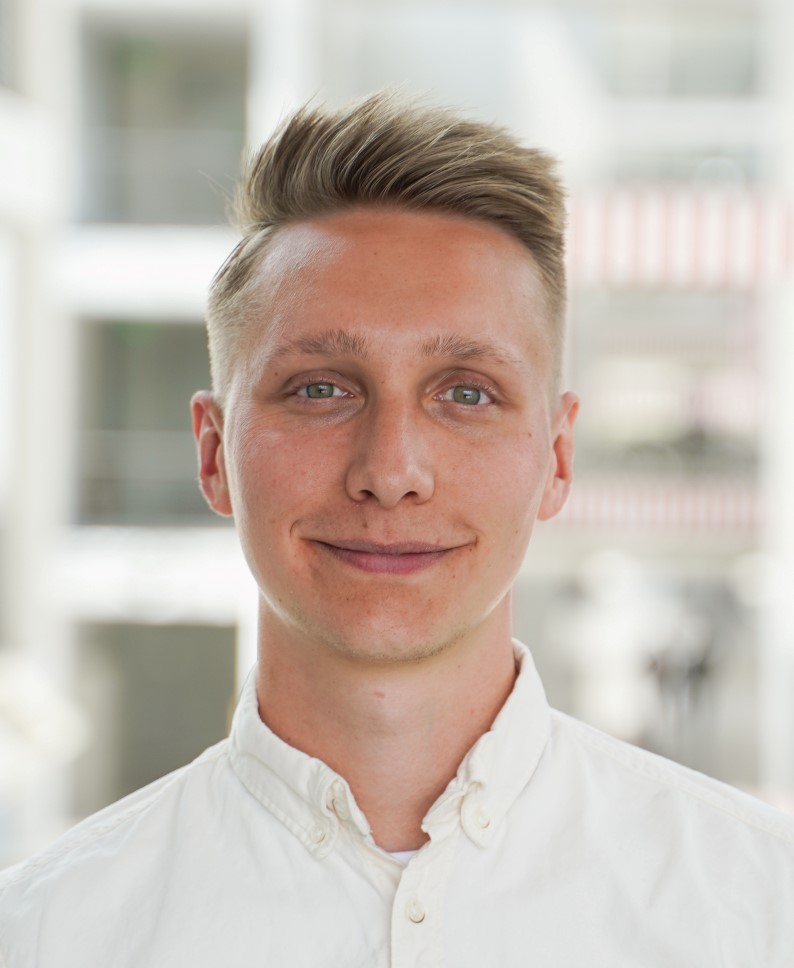}}]{Nico Uhlemann}
received his diploma from the Technical University of Dresden (TUD), Dresden, Germany, in 2020. After having worked in the industry as a system engineer, he joined the Institute of Automotive Technology at the Technical University of Munich in April of 2022 and is currently pursuing his Ph.D. there. His research interests include computer vision, hybrid prediction approaches and pedestrian intention estimation with respect to autonomous driving.
\end{IEEEbiography}

\begin{IEEEbiography}[{\includegraphics[width=1in,height=1.25in,clip,keepaspectratio]{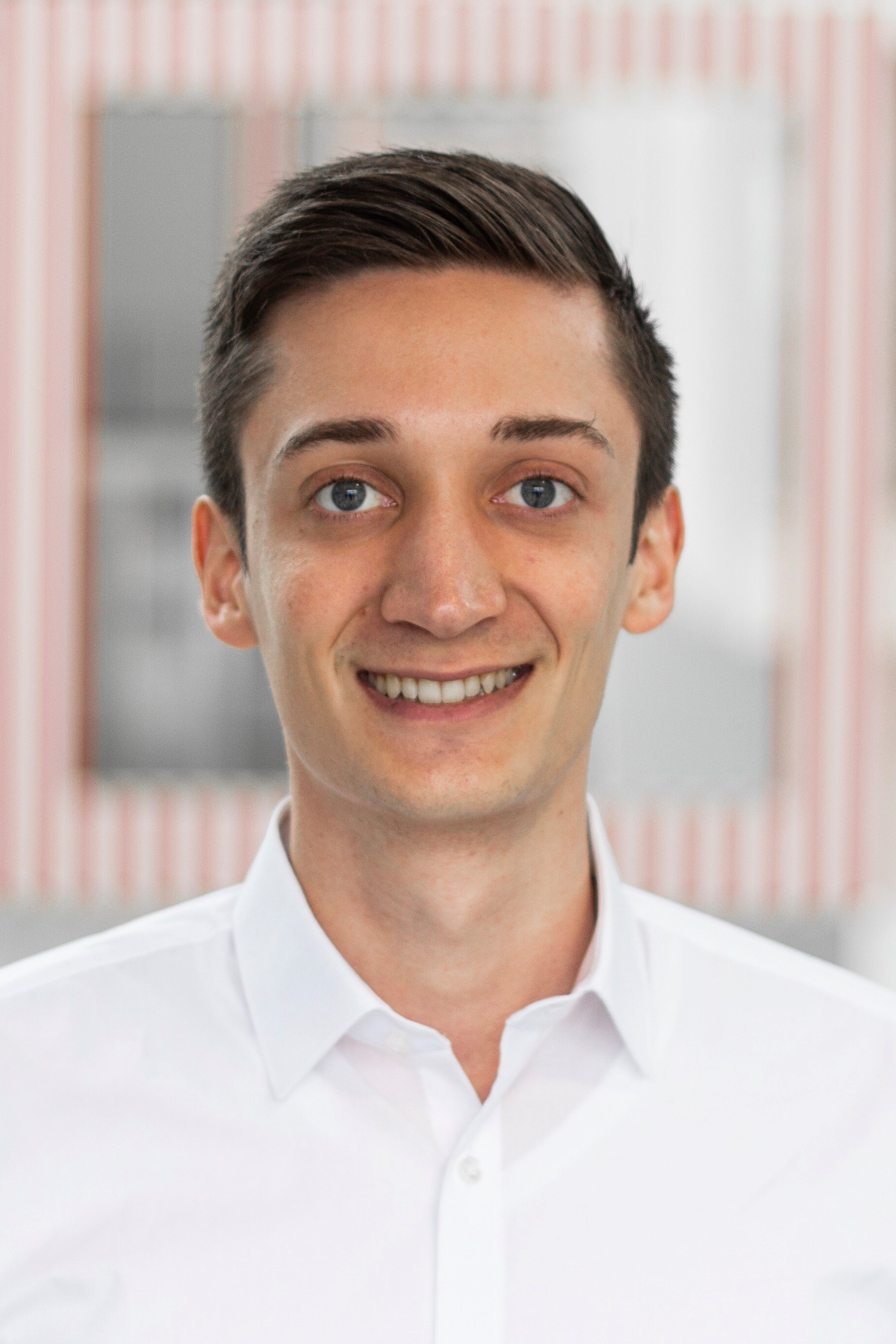}}]{Felix Fent}
received the B.Sc. and M.Sc. degrees from the Technical University of Munich (TUM), Munich, Germany, in 2018 and 2020, respectively, where he is currently pursuing a Ph.D. degree in mechanical engineering with the Institute of Automotive Technology. His research interests include radar-based perception, sensor fusion and multi-modal object detection approaches with a focus on real-world applications.
\end{IEEEbiography}

\begin{IEEEbiography}
[{\includegraphics[width=1in,height=1.25in,clip,keepaspectratio]{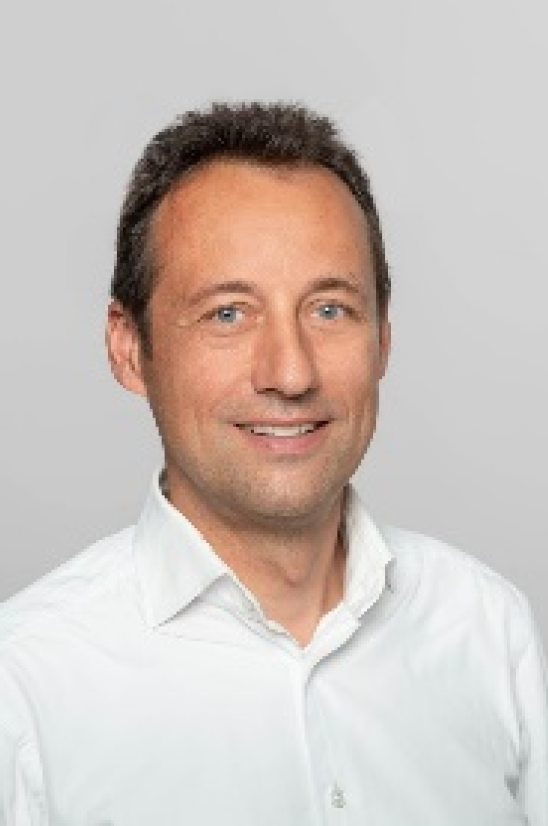}}]{Markus Lienkamp}
carries out research in the area of autonomous vehicles with the objective of creating an open-source software platform. He is a professor at the Institute of Automotive Technology at the Technical University of Munich (TUM). After studying mechanical engineering at TU Darmstadt and Cornell University, Prof. Lienkamp obtained his doctorate from TU Darmstadt (1995). He worked at Volkswagen as part of an international trainee program and took part in a joint venture between Ford and Volkswagen in Portugal. Returning to Germany, he led the brake testing department of the VW commercial vehicle development section in Wolfsburg. He was later appointed head of the Electronics and Vehicle research department in the Volkswagen Group’s Research Division. Prof. Lienkamp has headed the Chair of Automotive Technology at TUM since November 2009.
\end{IEEEbiography}

\end{document}

%% file: chapters/01_introduction.tex
\section{Introduction}
\label{sec_intro}

\begin{figure}[t]
    \centering
    \def\svgwidth{\linewidth}
    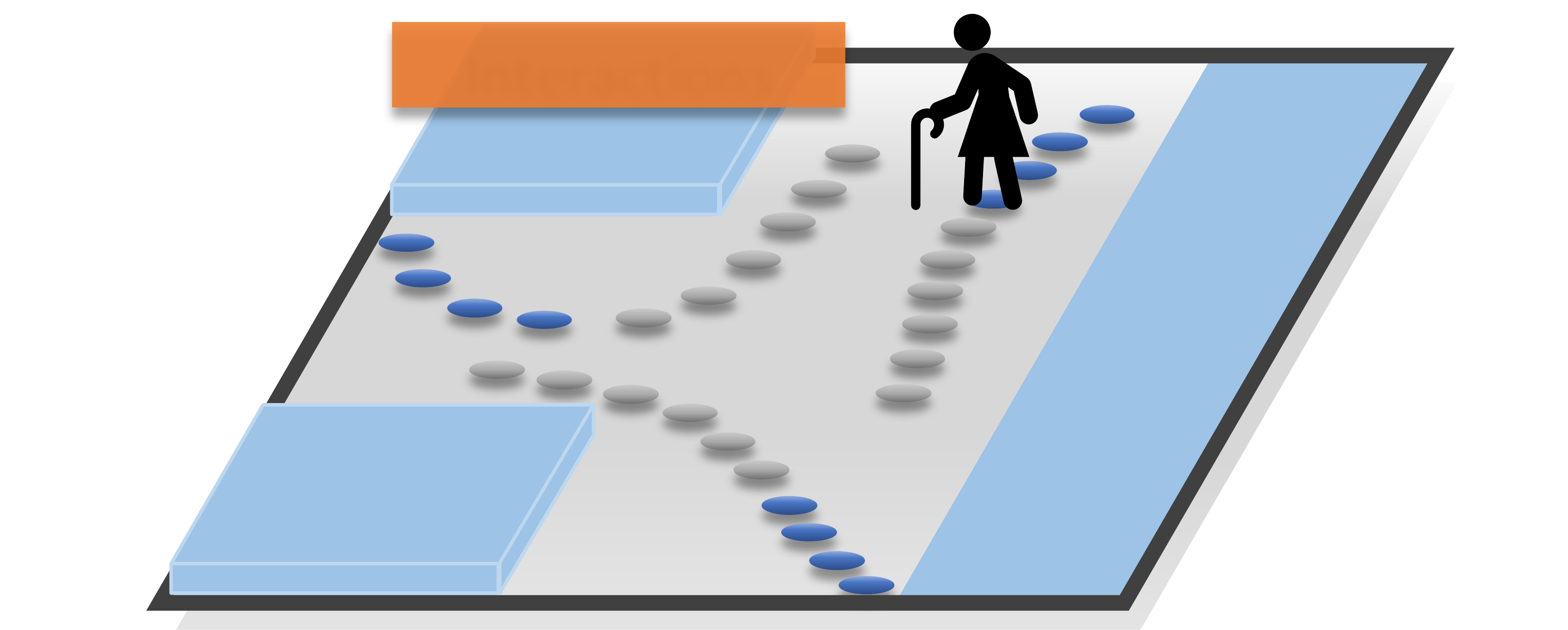
    \caption{The three features commonly employed for pedestrian trajectory prediction: the motion history of each agent, spatial information, and interactions.}
    \label{fig:pred_overview}
\end{figure}

\IEEEPARstart{A}{voiding} accidents involving vulnerable road users (VRU), such as pedestrians, is one of the paramount objectives and ongoing challenges for autonomous vehicles. According to the road safety report published by the World Health Organization (WHO) in 2018, pedestrians account for 23\% of the 1.35 million deaths globally caused by road traffic accidents \cite{who_report}. Consequently, the prediction of pedestrian behavior remains an active area of research as evidenced by recent publications \cite{mohamed_social-implicit_2022, yue_human_2022, zhang_forceformer_2023, fang_behavioral_2022, sharma_pedestrian_2022}. This line of research, known as pedestrian trajectory prediction, focuses on the modeling of their future trajectories and enables autonomous systems to consider their behavior \cite{rudenko_human_2020}. The trajectories are estimated with a bird’s-eye-view of the scene and the motion history of each pedestrian as visualized in figure~\ref{fig:pred_overview}. In addition, interactions can be explicitly considered. Initially modeled through simple rule-based methods \cite{helbing_social_1995}, various architectures based on neural networks \cite{gupta_social_2018, salzmann_trajectron_2021, mohamed_social-implicit_2022} as well as hybrid approaches \cite{yue_human_2022, zhang_forceformer_2023} have emerged in recent years, trying to improve the overall accuracy of the predicted paths on widely-used benchmarks like ETH/UCY \cite{lerner_crowds_2007, pellegrini_youll_2009} and SDD \cite{charalambous_learning_2016}. Although datasets tailored for traffic environments exist that are better suited for autonomous driving applications \cite{caesar_nuscenes_2020}, their evaluation predominantly revolves around vehicles. Hence, the development and assessment of pedestrian trajectory prediction methods primarily still occurs on the former. While over the years this development has undoubtedly resulted in significant advancements in the field, the adopted Best-of-N (BoN) evaluation approach falls short in quantifying their suitability for autonomous systems. In this regard, it is crucial to determine the accuracy of the most probable prediction, the extent to which the motion history influences its precision, and how effectively these methods scale as the number of observed pedestrians increases. \\

Therefore, in this paper we want to build upon the work outlined in previous publications \cite{mohamed_social-implicit_2022, scholler_what_2020, sun_human_nodate} by evaluating the state of the art with respect to the accuracy, feature requirements, and computational efficiency when generating single trajectories. The contributions of this work are therefore fourfold:
\begin{itemize}
    \item \textbf{First}, we evaluate the overall performance by reporting the ADE and FDE metrics when sampling a single trajectory. This is a critical measure for practical applications where usually only one prediction per agent is utilized to plan safe and collision-free paths.
    \item \textbf{Second}, we examine the sensitivity to input features by limiting the observed motion history to a maximum of two timesteps. While this resembles an extreme case, a prediction based on an initial observation is crucial in practical applications. Furthermore, deeper insights into the inner workings of each method can be gained.
    \item \textbf{Third}, we benchmark all contributions on a GPU while measuring runtimes to gauge how well each approach scales with an increasing number of agents.
    \item \textbf{Fourth}, by analyzing the generated predictions, shortcomings of the investigated approaches are derived and directions for future work provided.
\end{itemize}

The paper is structured as follows: In chapter \ref{sec_rel_work} we refer to related work already conducted in this field and evaluation methods employed. Following that, our overall evaluation procedure is introduced in chapter \ref{sec_meth}. Based on the findings outlined in chapter \ref{sec_res}, we discuss the results obtained in chapter \ref{sec_disc} and derive potential steps to guide future research and development in chapter \ref{sec_conc}.

%% file: images/feature_overview.pdf_tex
\begingroup%
  \makeatletter%
  \providecommand\color[2][]{%
    \errmessage{(Inkscape) Color is used for the text in Inkscape, but the package 'color.sty' is not loaded}%
    \renewcommand\color[2][]{}%
  }%
  \providecommand\transparent[1]{%
    \errmessage{(Inkscape) Transparency is used (non-zero) for the text in Inkscape, but the package 'transparent.sty' is not loaded}%
    \renewcommand\transparent[1]{}%
  }%
  \providecommand\rotatebox[2]{#2}%
  \newcommand*\fsize{\dimexpr\f@size pt\relax}%
  \newcommand*\lineheight[1]{\fontsize{\fsize}{#1\fsize}\selectfont}%
  \ifx\svgwidth\undefined%
    \setlength{\unitlength}{1556.25bp}%
    \ifx\svgscale\undefined%
      \relax%
    \else%
      \setlength{\unitlength}{\unitlength * \real{\svgscale}}%
    \fi%
  \else%
    \setlength{\unitlength}{\svgwidth}%
  \fi%
  \global\let\svgwidth\undefined%
  \global\let\svgscale\undefined%
  \makeatother%
  \begin{picture}(1,0.40144578)%
    \lineheight{1}%
    \setlength\tabcolsep{0pt}%
    \put(0,0){\includegraphics[width=\unitlength,page=1]{images/feature_overview.pdf}}%
    \put(0.29577831,0.34698795){\color[rgb]{1,1,1}\makebox(0,0)[lt]{\lineheight{1.25}\smash{\begin{tabular}[t]{l}\textbf{Interactions}\end{tabular}}}}%
    \put(0,0){\includegraphics[width=\unitlength,page=2]{images/feature_overview.pdf}}%
    \put(0.03739759,0.07951807){\color[rgb]{1,1,1}\makebox(0,0)[lt]{\lineheight{1.25}\smash{\begin{tabular}[t]{l}\textbf{Spatial}\end{tabular}}}}%
    \put(0.15888675,0.07951807){\color[rgb]{1,1,1}\makebox(0,0)[lt]{\lineheight{1.25}\smash{\begin{tabular}[t]{l}\textbf{information}\end{tabular}}}}%
    \put(0,0){\includegraphics[width=\unitlength,page=3]{images/feature_overview.pdf}}%
    \put(0.72115663,0.27084337){\color[rgb]{1,1,1}\makebox(0,0)[lt]{\lineheight{1.25}\smash{\begin{tabular}[t]{l}\textbf{Motion }\end{tabular}}}}%
    \put(0.85230843,0.27084337){\color[rgb]{1,1,1}\makebox(0,0)[lt]{\lineheight{1.25}\smash{\begin{tabular}[t]{l}\textbf{history}\end{tabular}}}}%
  \end{picture}%
\endgroup%

%% file: chapters/02_related_work.tex
\section{Related Work}
\label{sec_rel_work}

\begin{table*}[b]
\centering
\caption{Overview of employed model features and outputs on the ETH/UCY benchmark.}
\begin{tabular}{c|c||c|c|c|c|c|c}
\hline
 Model & Publ. & \multicolumn{2}{c|}{Motion history input (trajectories)} & Use of spatial & Interaction & \multicolumn{2}{c}{Output}\\
 & Year& Traj. considered & Format & information (map) & consideration & Number of traj. & Format \\
 \hline
 SGAN & 2018 & Multiple & Sequential & No & Capable & Multiple & Sequential \\
 Y-Net & 2021 & Multiple & Parallel & Yes & Yes (implicit) & Single & Parallel \\
 Trajectron++ & 2020 & Multiple & Sequential & Capable & Yes & Multiple & Parallel \\
 Social-Implicit & 2022 & Multiple & Parallel & No & Yes & Multiple & Parallel \\
 AgentFormer & 2021 & Multiple & Parallel & Capable & Yes & Multiple & Sequential \\
 CVM & - & Single & Sequential & No & No & Single & Sequential \\
 \hline
\end{tabular}
\label{table:overview_inputs}
\end{table*}

\textbf{Pedestrian trajectory prediction} can be categorized into two different approaches. The first category encompasses knowledge-based methods, with the simplest model being the constant velocity model (CVM) \cite{scholler_what_2020}. This model predicts linear trajectories based on an agent's most recent observation and is used as a baseline for trajectory prediction. A more sophisticated approach is the social force model \cite{helbing_social_1995}, which models human motion by considering attractive and repulsive forces that characterize the agent's interaction with the environment. It has found extensive usage in simulation environments for controlling interactive agents. Expanding on this model, advanced algorithms such as \textit{ORCA} \cite{van_den_berg_reciprocal_2011} and \textit{BRVO} \cite{kim_brvo_2015} have been developed and adapted for predicting human motion. These models consider the velocities of surrounding obstacles to estimate collision-free paths. The introduction of deep learning frameworks has led to a shift in focus, and the research community has employed neural networks to better approximate pedestrian trajectories. This second group of approaches is known as learning-based methods. In 2016, Social-LSTM \cite{alahi_social_2016} was published and has since remained one of the most influential approaches in the field. It utilizes a Recurrent Neural Network (RNN) to consider the past trajectory of observed agents and incorporates them into a pooling module to model interactions. Building upon the same architecture, Social GAN (SGAN) \cite{gupta_social_2018} was introduced in 2018, replacing the RNN architecture with a Generative Adversarial Network (GAN) and improving model accuracy by sampling reasonable and diverse trajectories. In 2020, Trajectron++ \cite{salzmann_trajectron_2021}, a Graph Neural Network (GNN), pushed the boundaries further. It leverages graph representations to account for cross-class interactions, such as between a car and a pedestrian. Currently, Y-Net \cite{mangalam_goals_2020} holds the second position in the common ETH/UCY benchmark \cite{alahi_social_2016} and, to the best of our knowledge, is the best-performing open-source contribution in this field. Y-Net utilizes an Encoder-Decoder setup, incorporating past trajectories and semantic maps to generate multimodal trajectories. AgentFormer \cite{yuan_agentformer_2021} was published shortly afterward, using a Transformer architecture with a specifically designed attention mechanism to model interactions. Since then, limited progress in enhancing the prediction accuracy has prompted researchers to explore other crucial factors of trajectory prediction methods. As a result, Social-Implicit \cite{mohamed_social-implicit_2022} was introduced in 2022, leveraging feed-forward neural networks in place of recurrent ones. This shift yielded enhancements in computational efficiency without compromising on the model's accuracy, thereby contributing to the progression in this field. \\

\textbf{The selection of models} for this evaluation was based on their strong performance, either recently or historically, establishing them as state of the art in various aspects. Each model represents a distinct neural network architectural type addressing the same problem, enabling a more insightful comparison. The chosen models include SGAN \cite{gupta_social_2018}, Trajectron++ \cite{salzmann_trajectron_2021}, AgentFormer \cite{yuan_agentformer_2021}, Y-Net \cite{mangalam_goals_2020}, and Social-Implicit \cite{mohamed_social-implicit_2022}. To establish a deterministic baseline, these models were benchmarked against a CVM. With the exception of the CVM, each architecture incorporates the motion history of multiple agents. SGAN and Trajectron++ sequentially process individual timesteps through their underlying RNN-based architecture, while the remaining ones process them in parallel. All models utilize positional data from the dataset with some also incorporating velocity (AgentFormer, Trajectron++, Social-Implicit) and, in the case of Trajectron++, even acceleration information. Additionally, explicit modeling of interactions with other agents in the scene is common among the investigated approaches, except for CVM and Y-Net. The CVM disregards interactions entirely, while Y-Net indirectly considers interactions by encoding the observed motion history in a heatmap. The mechanisms employed for modeling interactions directly are social pooling (SGAN), including spatial information within a spatial-temporal graph (Trajectron++, Social-Implicit), or the consideration in a specially designed attention layer (AgentFormer). The last feature used for the prediction is spatial information about the environment which is represented through simple obstacle maps or semantic maps containing roads, walkways, and non-walkable areas. Among these models, only Y-Net actively employs an obstacle map with two distinct classes. Trajectron++ and AgentFormer have the capability to include this type of information but have not utilized it in the given benchmark. Nevertheless, this marks them as the only two models to explicitly attempt the simultaneous integration of temporal, semantic, and interaction features. Regarding the output representation of the models, no clear trend is observed. While CVM, SGAN, and AgentFormer output timesteps sequentially, the remaining architectures generate trajectories in one pass over the entire prediction horizon (parallel). The results of this evaluation are summarized in TABLE~\ref{table:overview_inputs} wherein the term 'capable' indicates the model's ability to incorporate the stated information. However, such information remained unused in the ETH/UCY dataset based on the original implementation. \\

\textbf{To assess the effectiveness} of a given method, various metrics are commonly employed, with the Average Displacement Error and Final Displacement Error being the most prevalent ones \cite{rudenko_human_2020, sighencea_review_2021}. ADE measures the average Euclidean distance between predicted trajectories and observed ground truth trajectories, whereas FDE focuses solely on the last position, representing a measure of the error accumulation over the entire prediction horizon. In this evaluation, the models predict \SI{12}{} future timesteps based on the motion history derived from the last eight positional observations. To account for the multimodal behavior of humans, a Best-of-N (BoN) approach is typically used, sampling \SI{20}{} potential trajectories \cite{gupta_social_2018, salzmann_trajectron_2021, mangalam_goals_2020, yuan_agentformer_2021}. The trajectory with the smallest ADE is then selected to determine the overall performance. While this evaluation procedure has yielded significant progress in recent years, it has also faced criticism. Firstly, Schöller et al. \cite{scholler_what_2020} addressed the overall complexity of neural networks by demonstrating that a CVM can yield comparable results when the same BoN evaluation is applied to it. Secondly, concerns have been raised regarding the metric itself, as small error reductions during the evaluation on the ETH/UCY benchmark might also be caused by the inherent non-deterministic nature of Neural Networks \cite{mohamed_social-implicit_2022}. To address this, the authors propose a confidence-based metric for evaluating trajectory distributions. Lastly, the applicability of prediction algorithms in real-world scenarios, partially addressed in \cite{scholler_what_2020, sun_human_nodate}, remains a pressing concern. Consequently, the primary focus of this work is the evaluation of open-source state-of-the-art prediction models while generating a single trajectory for each pedestrian in the given scenario.

%% file: chapters/03_methodology.tex
\section{Methodology}
\label{sec_meth}

In terms of practical implementation for autonomous driving, the most significant factors of a prediction algorithm encompass the precision of the most likely trajectory, the prerequisite for an available trajectory from an initial observation, and the runtime required for generating said trajectories for the whole scenario. Consequently, the evaluation is performed across the following three criteria: Accuracy (BoN, most likely), feature requirements (limiting the number of temporal observations), and the inference time of each model. Whenever possible, the reported results were cross-checked against the values presented in the original publications or contributions found within the issues of the respective repositories. The following three sections outline our evaluation procedure, describe the data utilized for testing and training, and provide details about the explicit implementation of said experiments.

\subsection{Evaluation procedure}
\label{subsec:overall_eval}
To provide a measure for an overall comparison, the ADE and FDE were determined with the evaluation scripts provided by each individual contribution. Although these metrics don't provide a sufficient measure when a BoN evaluation is employed as has been previously addressed by Mohamed et al. \cite{mohamed_social-implicit_2022}, they provide valuable insight into the overall accuracy when single trajectories are predicted. Since the majority of models are non-deterministic and, in some cases, trajectories are randomly sampled rather than selecting the most likely one, each of the five scenes in the ETH/UCY dataset was evaluated five times. Afterwards, the average for each scene was calculated before combining all values to the overall results presented in figures \ref{fig:ade_fde_comp} and \ref{fig:feature_requ}. To sample a varying number of trajectories, the respective variables in each contribution were adjusted to either one or \SI{20}{}. Additional details are provided in \cref{subsec:impl_det}. 

As previously stated, almost all input features employed are present in real-world scenarios. Among the investigated approaches, Y-Net is the only model utilizing a semantic map on the given dataset and heavily relies on it due to its encoding mechanism. For this reason, an evaluation of the spatial information's influence on the model's performance was not conducted. Instead, the focus was shifted to the temporal information since all algorithms incorporate a motion history of eight timesteps for each agent within a scene. To investigate the impact of this feature, the number of available observations was limited and the corresponding metrics were computed. Specifically, the input tensor of each model was modified to contain either only the last timestep or the last two. Here, a timestep refers to an observation at any given time, encompassing both positional information and velocity data. While this limitation represents an extreme condition since not much information about an agent's movement can be derived, in practical applications a prediction might be needed right from the initial observation. Moreover, by evaluating these extreme cases, further insight into the workings of each method can be gained. Although experiments with more than two timesteps were conducted, these variations did not yield significant improvements and are therefore not reported in this study. For the most part, the missing entries were represented by a zero value within the respective tensor which applies to the positional data as well as the relative displacements and velocities. All other adjustments differing from this approach are listed in the implementation details below.

In order to evaluate the scalability of each approach as the number of observed agents increases, we measured the execution time of each model using the built-in \textit{CUDA}-API for PyTorch \cite{NEURIPS2019_9015}. This API offers precise time measurements through an accurate event recording. Specifically, we measured the execution time of a model starting from the point of providing individual scenario inputs until receiving predictions for all agents within the scene. The data preprocessing time was excluded from these measurements. To allow for a fair comparison, all experiments were performed on a \textit{Nvidia Tesla V100} device with \textit{CUDA} version 10.1 and Python 3.8.10. To ensure reliable estimates of the average scene prediction time, a batch size of one was chosen, and each model was evaluated five times. The median execution time and the interquartile range (IQR) were calculated across all scenes, providing insight into the range of runtimes for different numbers of agents. To visualize these results, Appendix~\ref{app:eval_runtime} contains a detailed overview of the two models that exhibited the largest variations across all employed scene sizes.

\subsection{Dataset and Training}
\label{subsec:data}
This study focuses on the assessment of short-term pedestrian trajectory prediction algorithms for time horizons up to \SI{4.8}{\s}. To facilitate an effective comparison and encompass scenarios of diverse complexity, the widely used ETH/UCY dataset \cite{lerner_crowds_2007, pellegrini_youll_2009} was selected. This dataset includes situations with over \SI{50}{} agents as well as scenarios with predominantly non-linear trajectories \cite{kothari_human_2022}, enabling an extensive evaluation of the chosen models. The data splits employed are the ones initially introduced by \cite{gupta_social_2018}. In the conducted Leave-One-Out Cross-Validation, four out of five scenes (Eth, Hotel, Univ, Zara1, and Zara2) were used for training and the remaining one for testing. In cases where the network weights were not provided alongside the code for the selected method, the models were trained according to the procedure outlined in each publication. An exception was made for Y-Net, as specific parameters were not given, and the provided code did not yield the expected results. 
Thus, a hyperparameter tuning was necessary, with the optimal parameter settings being 5.0 for \textit{Resize}, 0.5 for \textit{Temperature}, 31 for \textit{Kernel size}, and 4 for \textit{Sigma Gaussian Kernel}. Afterwards, the training was conducted for 150 epochs employing a batch size of 16, two semantic classes and a learning rate of 0.0001. Besides this, no retraining was done between the individual evaluation steps, adjusting only the necessary variables in the code to vary the number of generated trajectories.

\subsection{Implementation details}
\label{subsec:impl_det}
Alongside the steps described in the overall evaluation procedure, additional implementation details are provided in this section. \\

\textbf{SGAN}. During our evaluation as well as indicated in the paper itself \cite{gupta_social_2018}, the model without the pooling module delivered slightly better results, offering the same performance with less complexity. Therefore, results are reported for this configuration. \\
\textbf{Trajectron++}. The model provides different evaluation modes to assess the quality of the generated predictions. For this study, the Z-Mode was used over the 'most likely'-mode employed in the paper since it delivered almost identical results while decreasing the runtime drastically. \\
\textbf{Y-Net}. The default sampling of \SI{20}{} potential trajectories employs two modes, \textit{TTST} and \textit{CWS}. However, since clustering cannot be applied to a single trajectory, the \textit{TTST} mode was disabled, reducing inference time. To limit the observed motion history, missing timesteps were overwritten with the earliest observation since Y-Net uses positional data to encode the trajectory into the provided map.\\
\textbf{AgentFormer}. For the sampling of \SI{1}{} trajectory, the provided pre-trained models with the suffix \textit{\_pre} were used since the diverse sampling technique \textit{DLow} wasn't applicable. For the feature evaluation, non-existent entries are set to \SI{-1}{} \cite{kitti2012}. Within the attention matrix, the earliest observation was repeated for the remaining entries, delivering the best results.\\

%% file: chapters/04_results.tex
\section{Results}
\label{sec_res}
In this section, the results of the conducted evaluation are reported. In the first part, the general performance of each model is compared between the BoN approach and the most likely trajectory sampling. Subsequently, the second and third paragraphs focus on the temporal feature requirements and the execution time for predicting an entire scenario. Within this section, the variable $K$ refers to the number of trajectories sampled whereas $N$ stands for the number of timesteps provided to the network.

\subsection{Accuracy}

While some methods benchmark the performance against a linear regression model as a baseline \cite{gupta_social_2018, salzmann_trajectron_2021}, fitting a straight line to the observed motion history, a CVM has not been utilized as a reference in the investigated contributions before. However, due to its proven performance in previous studies \cite{scholler_what_2020}, the CVM was selected as a baseline for this evaluation. The average values of the ADE and FDE for one and \SI{20}{} generated future trajectories are presented in Fig.~\ref{fig:ade_fde_comp}, with the exact values for each scene being contained in Appendix~\ref{app:eval_values}. The input configuration for each model is described in TABLE~\ref{table:overview_inputs}, where the motion history consists of eight and the prediction horizon of \SI{12}{} timesteps.

\begin{figure}[t]
    \centering
    \fontsize{8pt}{12pt}\selectfont
    \def\svgwidth{0.96\linewidth}
    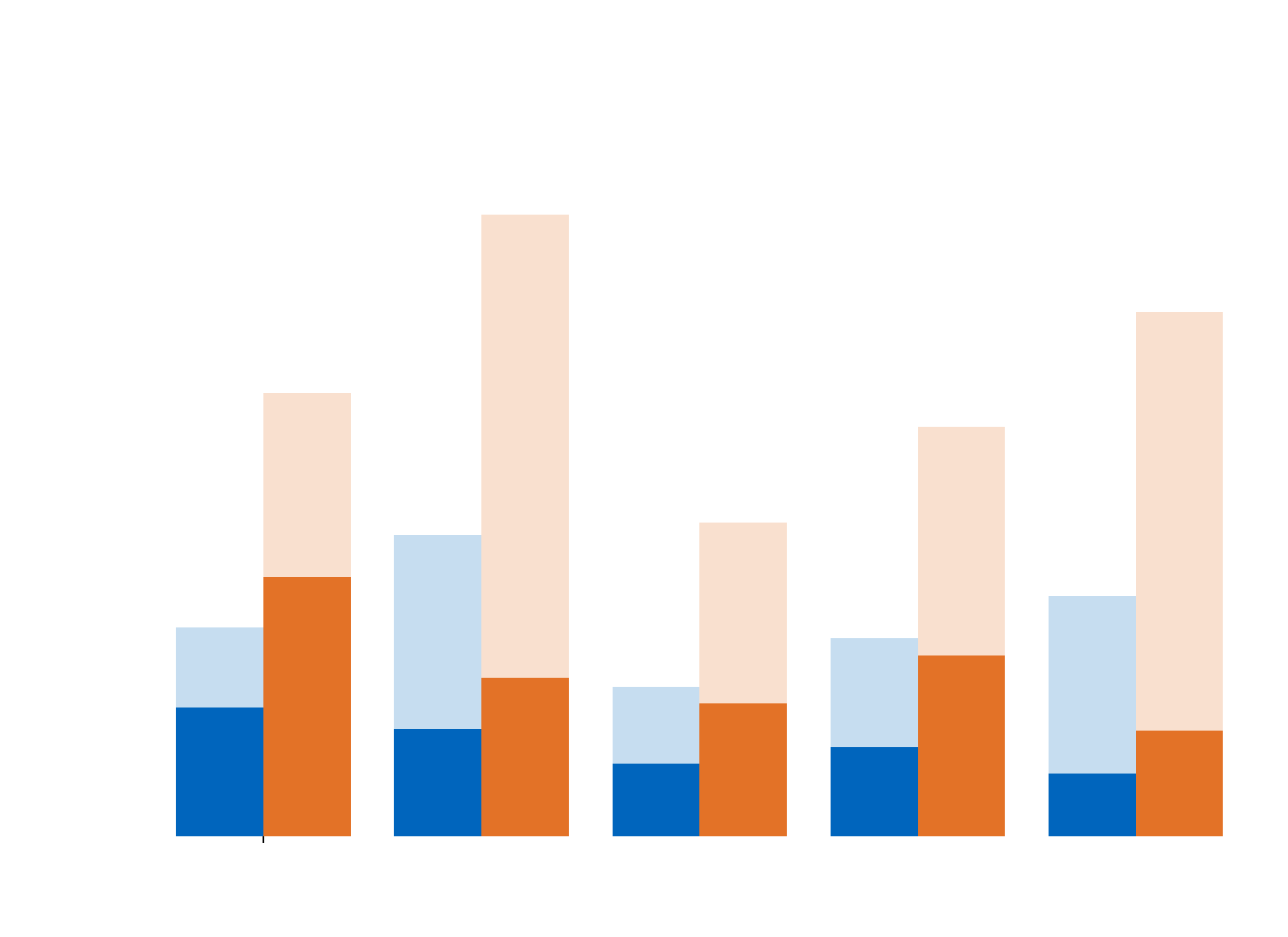
    \caption{ADE (blue) and FDE (orange) across various models using one and \SI{20}{} sampled trajectories}
    \label{fig:ade_fde_comp}
\end{figure}

In Fig.~\ref{fig:ade_fde_comp}, blue shades represent the ADE metric and orange ones highlight the FDE scores. Lighter colors were chosen for the evaluation of a single trajectory whereas darker ones visualize the results obtained for the BoN approach, evaluating the best of $K=20$ sampled trajectories against the ground truth. The horizontal, dashed lines indicate the results of the CVM, following the same color scheme. When sampling \SI{20}{} distinct trajectories, the investigated models demonstrate the anticipated strong performance on the ETH/UCY dataset, outperforming the simple CVM baseline with an ADE of \SI{0.52}{\meter} and an FDE of \SI{1.141}{\meter}. However, we were unable to replicate the reported results of Y-Net despite extensive parameter tuning, and the values for Trajectron++ are slightly worse due to previously reported issues with the velocity and acceleration derivatives. Taking these factors into account, AgentFormer achieves the best performance in our BoN evaluation with \SI{0.233}{\meter} and \SI{0.393}{\meter}, respectively. With an increase in the average displacement of \SI{3.8}{\centi\meter} and \SI{10.1}{\centi\meter} for the FDE, Trajectron++ scores second. This changes when sampling only one trajectory and selecting the most probable one whenever possible. In this configuration, while all Final Displacement Errors now exceed one meter, the SGAN architecture is among the least affected and delivers competitive results. Conversely, AgentFormer exhibits a significant decline in performance, ranking second last in the provided overview. Trajectron++ yields the most precise predictions with \SI{0.555}{\meter} in average and \SI{1.162}{\meter} in final displacement. Interestingly, none of the featured models surpass the performance of the CVM which only considers information from the last timestep. Therefore, the influence of individual timesteps on the performance of each model will be examined in the following section.

\subsection{Feature requirements}
\begin{figure}[t]
    \centering
    \fontsize{8pt}{12pt}\selectfont
    \def\svgwidth{\linewidth}
    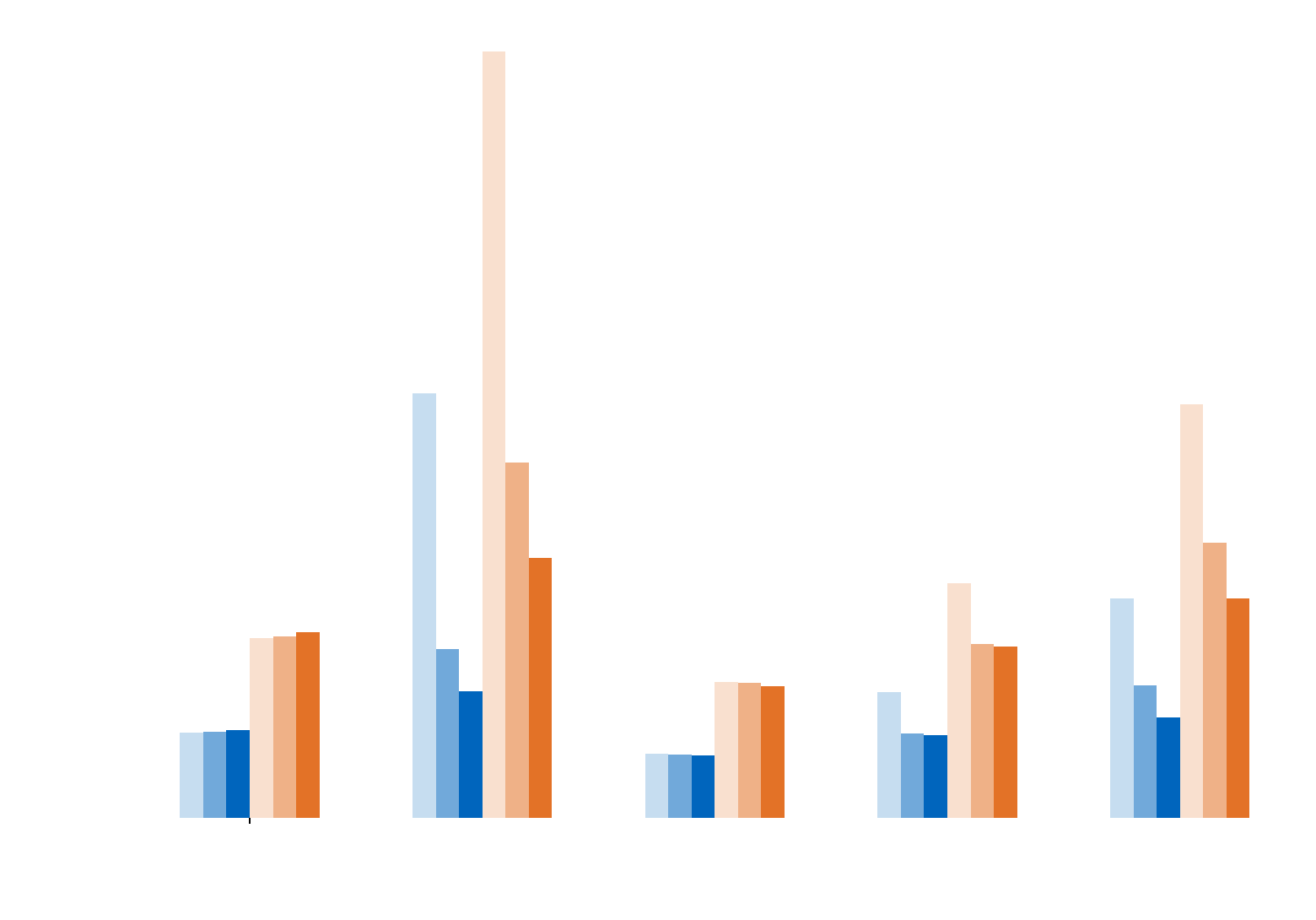
    \caption{ADE (blue) and FDE (orange) across various models for $N = {1,2,8}$ observations}
    \label{fig:feature_requ}
\end{figure}

Previous studies suggest that the motion history of an agent may not be as crucial as commonly assumed \cite{scholler_what_2020}. Therefore, in order to gain insights into the inner workings of each method, we examined the influence of one, two, and eight observed timesteps provided to each model. Details on the necessary modifications for this analysis can be found in section \ref{sec_meth}, and the summarized results are presented in figure~\ref{fig:feature_requ}.

Similar to the previous evaluation, the figure highlights ADE values in blue whereas FDE values are represented with shades of orange. The brighter the tone, the fewer observations were provided to the network. Again, the respective metrics corresponding to the performance of the CVM are indicated through dashed, horizontal lines. When focusing on the overall trend between the metrics, it can be seen that first, both metrics express a similar behavior across all investigated methods, and second, the lowest errors are achieved when all observations are considered. Reducing the amount of information available to one timestep, not all models become worse to the same degree. Whereas AgentFormer and Y-Net exhibit the worst performance in our evaluation with Average Displacement Errors up to \SI{1.941}{\meter} and \SI{3.757}{\meter} respectively, the remaining three models show almost no difference when provided with only one or two timesteps. While the ADE for Social-Implicit improves from \SI{1.11}{\meter} to \SI{0.745}{\meter} when adding an additional observation, the change in accuracy for Trajectron++ and SGAN is less than \SI{1}{\centi\meter}. These displacements are almost identical to the ones obtained when considering the whole motion history with gains being as little as \SI{0.4}{\centi\meter} in the case of Trajectron++. However, when an additional timestep is provided to both Y-Net and AgentFormer, a substantial decrease in both Average and Final Displacement Errors can be noted, with differences as high as \SI{2.261}{\meter} and \SI{3.635}{\meter}, respectively. Further improvements in accuracy can be observed when considering the full motion history, although the increase is significantly smaller in comparison.

\subsection{Runtime}
\begin{figure}[t]
    \centering
    \fontsize{8pt}{12pt}\selectfont
    \def\svgwidth{\linewidth}
    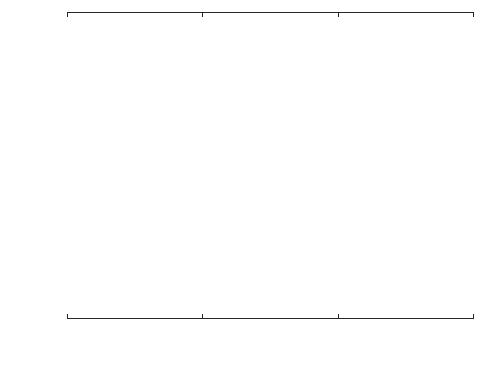
    \caption{Runtime of the pedestrian trajectory prediction models represented by their median inference time over all scenarios and their variability as interquartile range (IQR) between the 90th and 10th percentiles over the Average Displacement Error for $K=1$ and $N=8$.}
    \label{fig:runtime}
\end{figure}

The runtime analysis shows a significant difference in the inference time of the compared models as visualized in figure~\ref{fig:runtime}. According to this, the CVM has the lowest runtime with a median inference time of \SI{0.15}{\ms}, while the highest median inference time was measured for Trajectron++ with \SI{131.34}{\ms}. However, not just the central tendency but also the runtime variability is an important measure for real-time applications. As a result, the CVM has the smallest interquartile range with a deviation of \SI{0.01}{\ms} between the 90th and 10th percentiles. In contrast, Trajectron++ has the highest variability with an IQR of \SI{197.54}{\ms}. For the remaining models, the median runtime ranges between \SI{1.69}{\ms}, \SI{4.02}{\ms} and \SI{49.01}{\ms} for Social-Implicit, SGAN and AgentFormer, respectively. Furthermore, it can be observed that both Social-Implicit and SGAN have a low variability with an IQR of \SI{0.11}{\ms} and \SI{0.22}{\ms}, and therefore indicating a weak runtime dependency on the number of agents in the scene. In contrast, Y-Net has a median runtime of \SI{82.39}{\ms} and shows a large variability with an IQR of \SI{96.87}{\ms}, similar to Trajectron++. In comparison, AgentFormer has an IQR of \SI{4.92}{\ms}. An evaluation of the relationship between the runtime and the number of agents in a scene is provided in Appendix~\ref{app:eval_runtime}.

The results also indicate no direct correlation between the model runtime and the achieved ADE. Even if the CVM has both the lowest runtime and the smallest ADE, the second most accurate network (Trajectron++) has the highest median runtime overall. Moreover, both Social-Implicit as well as SGAN achieve comparable accuracies, but their inference time deviates by more than an order of magnitude from the runtime of the CVM. While this observation holds true for AgentFormer, the Y-Net model has the highest ADE, but only the third-highest median runtime.

%% file: images/ade_fde_comp_2.pdf_tex
\begingroup%
  \makeatletter%
  \providecommand\color[2][]{%
    \errmessage{(Inkscape) Color is used for the text in Inkscape, but the package 'color.sty' is not loaded}%
    \renewcommand\color[2][]{}%
  }%
  \providecommand\transparent[1]{%
    \errmessage{(Inkscape) Transparency is used (non-zero) for the text in Inkscape, but the package 'transparent.sty' is not loaded}%
    \renewcommand\transparent[1]{}%
  }%
  \providecommand\rotatebox[2]{#2}%
  \newcommand*\fsize{\dimexpr\f@size pt\relax}%
  \newcommand*\lineheight[1]{\fontsize{\fsize}{#1\fsize}\selectfont}%
  \ifx\svgwidth\undefined%
    \setlength{\unitlength}{748.71876526bp}%
    \ifx\svgscale\undefined%
      \relax%
    \else%
      \setlength{\unitlength}{\unitlength * \real{\svgscale}}%
    \fi%
  \else%
    \setlength{\unitlength}{\svgwidth}%
  \fi%
  \global\let\svgwidth\undefined%
  \global\let\svgscale\undefined%
  \makeatother%
  \begin{picture}(1,0.72719351)%
    \lineheight{1}%
    \setlength\tabcolsep{0pt}%
    \put(0,0){\includegraphics[width=\unitlength,page=1]{images/ade_fde_comp_2.pdf}}%
    \put(0.1646098,0.04687172){\color[rgb]{0,0,0}\makebox(0,0)[lt]{\lineheight{1.25}\smash{\begin{tabular}[t]{l}SGAN\end{tabular}}}}%
    \put(0,0){\includegraphics[width=\unitlength,page=2]{images/ade_fde_comp_2.pdf}}%
    \put(0.33726649,0.04687172){\color[rgb]{0,0,0}\makebox(0,0)[lt]{\lineheight{1.25}\smash{\begin{tabular}[t]{l}Y-Net\end{tabular}}}}%
    \put(0,0){\includegraphics[width=\unitlength,page=3]{images/ade_fde_comp_2.pdf}}%
    \put(0.49707831,0.04687172){\color[rgb]{0,0,0}\makebox(0,0)[lt]{\lineheight{1.25}\smash{\begin{tabular}[t]{l}Trajec-\end{tabular}}}}%
    \put(0.49131847,0.01545973){\color[rgb]{0,0,0}\makebox(0,0)[lt]{\lineheight{1.25}\smash{\begin{tabular}[t]{l}tron++\end{tabular}}}}%
    \put(0,0){\includegraphics[width=\unitlength,page=4]{images/ade_fde_comp_2.pdf}}%
    \put(0.66596816,0.04687172){\color[rgb]{0,0,0}\makebox(0,0)[lt]{\lineheight{1.25}\smash{\begin{tabular}[t]{l}Social-\end{tabular}}}}%
    \put(0.6610222,0.01545973){\color[rgb]{0,0,0}\makebox(0,0)[lt]{\lineheight{1.25}\smash{\begin{tabular}[t]{l}Implicit\end{tabular}}}}%
    \put(0,0){\includegraphics[width=\unitlength,page=5]{images/ade_fde_comp_2.pdf}}%
    \put(0.83541103,0.04687172){\color[rgb]{0,0,0}\makebox(0,0)[lt]{\lineheight{1.25}\smash{\begin{tabular}[t]{l}Agent-\end{tabular}}}}%
    \put(0.83225981,0.01545973){\color[rgb]{0,0,0}\makebox(0,0)[lt]{\lineheight{1.25}\smash{\begin{tabular}[t]{l}Former\end{tabular}}}}%
    \put(0,0){\includegraphics[width=\unitlength,page=6]{images/ade_fde_comp_2.pdf}}%
    \put(0.04210944,0.06687464){\color[rgb]{0,0,0}\makebox(0,0)[lt]{\lineheight{1.25}\smash{\begin{tabular}[t]{l}0.0\end{tabular}}}}%
    \put(0,0){\includegraphics[width=\unitlength,page=7]{images/ade_fde_comp_2.pdf}}%
    \put(0.04210944,0.17177384){\color[rgb]{0,0,0}\makebox(0,0)[lt]{\lineheight{1.25}\smash{\begin{tabular}[t]{l}0.5\end{tabular}}}}%
    \put(0,0){\includegraphics[width=\unitlength,page=8]{images/ade_fde_comp_2.pdf}}%
    \put(0.04210944,0.27667303){\color[rgb]{0,0,0}\makebox(0,0)[lt]{\lineheight{1.25}\smash{\begin{tabular}[t]{l}1.0\end{tabular}}}}%
    \put(0,0){\includegraphics[width=\unitlength,page=9]{images/ade_fde_comp_2.pdf}}%
    \put(0.04210944,0.38157223){\color[rgb]{0,0,0}\makebox(0,0)[lt]{\lineheight{1.25}\smash{\begin{tabular}[t]{l}1.5\end{tabular}}}}%
    \put(0,0){\includegraphics[width=\unitlength,page=10]{images/ade_fde_comp_2.pdf}}%
    \put(0.04210944,0.48647143){\color[rgb]{0,0,0}\makebox(0,0)[lt]{\lineheight{1.25}\smash{\begin{tabular}[t]{l}2.0\end{tabular}}}}%
    \put(0,0){\includegraphics[width=\unitlength,page=11]{images/ade_fde_comp_2.pdf}}%
    \put(0.04210944,0.59137063){\color[rgb]{0,0,0}\makebox(0,0)[lt]{\lineheight{1.25}\smash{\begin{tabular}[t]{l}2.5\end{tabular}}}}%
    \put(0,0){\includegraphics[width=\unitlength,page=12]{images/ade_fde_comp_2.pdf}}%
    \put(0.04210944,0.69626983){\color[rgb]{0,0,0}\makebox(0,0)[lt]{\lineheight{1.25}\smash{\begin{tabular}[t]{l}3.0\end{tabular}}}}%
    \put(0.03092366,0.21453689){\color[rgb]{0,0,0}\rotatebox{90}{\makebox(0,0)[lt]{\lineheight{1.25}\smash{\begin{tabular}[t]{l}ADE and FDE values in m\end{tabular}}}}}%
    \put(0,0){\includegraphics[width=\unitlength,page=13]{images/ade_fde_comp_2.pdf}}%
    \put(0.55514419,0.66037309){\color[rgb]{0,0,0}\makebox(0,0)[lt]{\lineheight{1.25}\smash{\begin{tabular}[t]{l}ADE CVM\end{tabular}}}}%
    \put(0,0){\includegraphics[width=\unitlength,page=14]{images/ade_fde_comp_2.pdf}}%
    \put(0.55514419,0.61919859){\color[rgb]{0,0,0}\makebox(0,0)[lt]{\lineheight{1.25}\smash{\begin{tabular}[t]{l}FDE CVM\end{tabular}}}}%
    \put(0,0){\includegraphics[width=\unitlength,page=15]{images/ade_fde_comp_2.pdf}}%
    \put(0.55514419,0.57802408){\color[rgb]{0,0,0}\makebox(0,0)[lt]{\lineheight{1.25}\smash{\begin{tabular}[t]{l}ADE K=1\end{tabular}}}}%
    \put(0,0){\includegraphics[width=\unitlength,page=16]{images/ade_fde_comp_2.pdf}}%
    \put(0.82016358,0.66037309){\color[rgb]{0,0,0}\makebox(0,0)[lt]{\lineheight{1.25}\smash{\begin{tabular}[t]{l}FDE K=1\end{tabular}}}}%
    \put(0,0){\includegraphics[width=\unitlength,page=17]{images/ade_fde_comp_2.pdf}}%
    \put(0.82016358,0.61919859){\color[rgb]{0,0,0}\makebox(0,0)[lt]{\lineheight{1.25}\smash{\begin{tabular}[t]{l}ADE K=20\end{tabular}}}}%
    \put(0,0){\includegraphics[width=\unitlength,page=18]{images/ade_fde_comp_2.pdf}}%
    \put(0.82016358,0.57802408){\color[rgb]{0,0,0}\makebox(0,0)[lt]{\lineheight{1.25}\smash{\begin{tabular}[t]{l}FDE K=20\end{tabular}}}}%
  \end{picture}%
\endgroup%

%% file: images/feature_comp.pdf_tex
\begingroup%
  \makeatletter%
  \providecommand\color[2][]{%
    \errmessage{(Inkscape) Color is used for the text in Inkscape, but the package 'color.sty' is not loaded}%
    \renewcommand\color[2][]{}%
  }%
  \providecommand\transparent[1]{%
    \errmessage{(Inkscape) Transparency is used (non-zero) for the text in Inkscape, but the package 'transparent.sty' is not loaded}%
    \renewcommand\transparent[1]{}%
  }%
  \providecommand\rotatebox[2]{#2}%
  \newcommand*\fsize{\dimexpr\f@size pt\relax}%
  \newcommand*\lineheight[1]{\fontsize{\fsize}{#1\fsize}\selectfont}%
  \ifx\svgwidth\undefined%
    \setlength{\unitlength}{748.71876526bp}%
    \ifx\svgscale\undefined%
      \relax%
    \else%
      \setlength{\unitlength}{\unitlength * \real{\svgscale}}%
    \fi%
  \else%
    \setlength{\unitlength}{\svgwidth}%
  \fi%
  \global\let\svgwidth\undefined%
  \global\let\svgscale\undefined%
  \makeatother%
  \begin{picture}(1,0.69873639)%
    \lineheight{1}%
    \setlength\tabcolsep{0pt}%
    \put(0,0){\includegraphics[width=\unitlength,page=1]{images/feature_comp.pdf}}%
    \put(0.14988108,0.04687174){\color[rgb]{0,0,0}\makebox(0,0)[lt]{\lineheight{1.25}\smash{\begin{tabular}[t]{l}SGAN\end{tabular}}}}%
    \put(0,0){\includegraphics[width=\unitlength,page=2]{images/feature_comp.pdf}}%
    \put(0.32990213,0.04687174){\color[rgb]{0,0,0}\makebox(0,0)[lt]{\lineheight{1.25}\smash{\begin{tabular}[t]{l}Y-Net\end{tabular}}}}%
    \put(0,0){\includegraphics[width=\unitlength,page=3]{images/feature_comp.pdf}}%
    \put(0.49707831,0.04687174){\color[rgb]{0,0,0}\makebox(0,0)[lt]{\lineheight{1.25}\smash{\begin{tabular}[t]{l}Trajec-\end{tabular}}}}%
    \put(0.49131847,0.01545975){\color[rgb]{0,0,0}\makebox(0,0)[lt]{\lineheight{1.25}\smash{\begin{tabular}[t]{l}tron++\end{tabular}}}}%
    \put(0,0){\includegraphics[width=\unitlength,page=4]{images/feature_comp.pdf}}%
    \put(0.67333251,0.04687174){\color[rgb]{0,0,0}\makebox(0,0)[lt]{\lineheight{1.25}\smash{\begin{tabular}[t]{l}Social-\end{tabular}}}}%
    \put(0.66838657,0.01545975){\color[rgb]{0,0,0}\makebox(0,0)[lt]{\lineheight{1.25}\smash{\begin{tabular}[t]{l}Implicit\end{tabular}}}}%
    \put(0,0){\includegraphics[width=\unitlength,page=5]{images/feature_comp.pdf}}%
    \put(0.85013974,0.04687174){\color[rgb]{0,0,0}\makebox(0,0)[lt]{\lineheight{1.25}\smash{\begin{tabular}[t]{l}Agent-\end{tabular}}}}%
    \put(0.84698853,0.01545975){\color[rgb]{0,0,0}\makebox(0,0)[lt]{\lineheight{1.25}\smash{\begin{tabular}[t]{l}Former\end{tabular}}}}%
    \put(0,0){\includegraphics[width=\unitlength,page=6]{images/feature_comp.pdf}}%
    \put(0.04210944,0.06687466){\color[rgb]{0,0,0}\makebox(0,0)[lt]{\lineheight{1.25}\smash{\begin{tabular}[t]{l}0.0\end{tabular}}}}%
    \put(0,0){\includegraphics[width=\unitlength,page=7]{images/feature_comp.pdf}}%
    \put(0.04210944,0.15272295){\color[rgb]{0,0,0}\makebox(0,0)[lt]{\lineheight{1.25}\smash{\begin{tabular}[t]{l}1.0\end{tabular}}}}%
    \put(0,0){\includegraphics[width=\unitlength,page=8]{images/feature_comp.pdf}}%
    \put(0.04210944,0.23857125){\color[rgb]{0,0,0}\makebox(0,0)[lt]{\lineheight{1.25}\smash{\begin{tabular}[t]{l}2.0\end{tabular}}}}%
    \put(0,0){\includegraphics[width=\unitlength,page=9]{images/feature_comp.pdf}}%
    \put(0.04210944,0.32441954){\color[rgb]{0,0,0}\makebox(0,0)[lt]{\lineheight{1.25}\smash{\begin{tabular}[t]{l}3.0\end{tabular}}}}%
    \put(0,0){\includegraphics[width=\unitlength,page=10]{images/feature_comp.pdf}}%
    \put(0.04210944,0.41026783){\color[rgb]{0,0,0}\makebox(0,0)[lt]{\lineheight{1.25}\smash{\begin{tabular}[t]{l}4.0\end{tabular}}}}%
    \put(0,0){\includegraphics[width=\unitlength,page=11]{images/feature_comp.pdf}}%
    \put(0.04210944,0.49611611){\color[rgb]{0,0,0}\makebox(0,0)[lt]{\lineheight{1.25}\smash{\begin{tabular}[t]{l}5.0\end{tabular}}}}%
    \put(0,0){\includegraphics[width=\unitlength,page=12]{images/feature_comp.pdf}}%
    \put(0.04210944,0.58196441){\color[rgb]{0,0,0}\makebox(0,0)[lt]{\lineheight{1.25}\smash{\begin{tabular}[t]{l}6.0\end{tabular}}}}%
    \put(0,0){\includegraphics[width=\unitlength,page=13]{images/feature_comp.pdf}}%
    \put(0.04210944,0.6678127){\color[rgb]{0,0,0}\makebox(0,0)[lt]{\lineheight{1.25}\smash{\begin{tabular}[t]{l}7.0\end{tabular}}}}%
    \put(0.03092366,0.2052811){\color[rgb]{0,0,0}\rotatebox{90}{\makebox(0,0)[lt]{\lineheight{1.25}\smash{\begin{tabular}[t]{l}ADE and FDE values in m\end{tabular}}}}}%
    \put(0,0){\includegraphics[width=\unitlength,page=14]{images/feature_comp.pdf}}%
    \put(0.57039941,0.64186149){\color[rgb]{0,0,0}\makebox(0,0)[lt]{\lineheight{1.25}\smash{\begin{tabular}[t]{l}ADE CVM\end{tabular}}}}%
    \put(0,0){\includegraphics[width=\unitlength,page=15]{images/feature_comp.pdf}}%
    \put(0.57039941,0.60068699){\color[rgb]{0,0,0}\makebox(0,0)[lt]{\lineheight{1.25}\smash{\begin{tabular}[t]{l}FDE CVM\end{tabular}}}}%
    \put(0,0){\includegraphics[width=\unitlength,page=16]{images/feature_comp.pdf}}%
    \put(0.57039941,0.55951248){\color[rgb]{0,0,0}\makebox(0,0)[lt]{\lineheight{1.25}\smash{\begin{tabular}[t]{l}ADE N=1\end{tabular}}}}%
    \put(0,0){\includegraphics[width=\unitlength,page=17]{images/feature_comp.pdf}}%
    \put(0.57039941,0.51833797){\color[rgb]{0,0,0}\makebox(0,0)[lt]{\lineheight{1.25}\smash{\begin{tabular}[t]{l}ADE N=2\end{tabular}}}}%
    \put(0,0){\includegraphics[width=\unitlength,page=18]{images/feature_comp.pdf}}%
    \put(0.8354188,0.64186149){\color[rgb]{0,0,0}\makebox(0,0)[lt]{\lineheight{1.25}\smash{\begin{tabular}[t]{l}ADE N=8\end{tabular}}}}%
    \put(0,0){\includegraphics[width=\unitlength,page=19]{images/feature_comp.pdf}}%
    \put(0.8354188,0.60068699){\color[rgb]{0,0,0}\makebox(0,0)[lt]{\lineheight{1.25}\smash{\begin{tabular}[t]{l}FDE N=1\end{tabular}}}}%
    \put(0,0){\includegraphics[width=\unitlength,page=20]{images/feature_comp.pdf}}%
    \put(0.8354188,0.55951248){\color[rgb]{0,0,0}\makebox(0,0)[lt]{\lineheight{1.25}\smash{\begin{tabular}[t]{l}FDE N=2\end{tabular}}}}%
    \put(0,0){\includegraphics[width=\unitlength,page=21]{images/feature_comp.pdf}}%
    \put(0.8354188,0.51833797){\color[rgb]{0,0,0}\makebox(0,0)[lt]{\lineheight{1.25}\smash{\begin{tabular}[t]{l}FDE N=8\end{tabular}}}}%
  \end{picture}%
\endgroup%

%% file: images/Nico_Fig_Runtime_over_ADE_v02.pdf_tex
\begingroup%
  \makeatletter%
  \providecommand\color[2][]{%
    \errmessage{(Inkscape) Color is used for the text in Inkscape, but the package 'color.sty' is not loaded}%
    \renewcommand\color[2][]{}%
  }%
  \providecommand\transparent[1]{%
    \errmessage{(Inkscape) Transparency is used (non-zero) for the text in Inkscape, but the package 'transparent.sty' is not loaded}%
    \renewcommand\transparent[1]{}%
  }%
  \providecommand\rotatebox[2]{#2}%
  \newcommand*\fsize{\dimexpr\f@size pt\relax}%
  \newcommand*\lineheight[1]{\fontsize{\fsize}{#1\fsize}\selectfont}%
  \ifx\svgwidth\undefined%
    \setlength{\unitlength}{233.99999415bp}%
    \ifx\svgscale\undefined%
      \relax%
    \else%
      \setlength{\unitlength}{\unitlength * \real{\svgscale}}%
    \fi%
  \else%
    \setlength{\unitlength}{\svgwidth}%
  \fi%
  \global\let\svgwidth\undefined%
  \global\let\svgscale\undefined%
  \makeatother%
  \begin{picture}(1,0.76923077)%
    \lineheight{1}%
    \setlength\tabcolsep{0pt}%
    \put(0,0){\includegraphics[width=\unitlength,page=1]{images/Nico_Fig_Runtime_over_ADE_v02.pdf}}%
    \put(0.12660256,0.06495726){\color[rgb]{0.14901961,0.14901961,0.14901961}\makebox(0,0)[lt]{\lineheight{1.25}\smash{\begin{tabular}[t]{l}0\end{tabular}}}}%
    \put(0.3883547,0.06495726){\color[rgb]{0.14901961,0.14901961,0.14901961}\makebox(0,0)[lt]{\lineheight{1.25}\smash{\begin{tabular}[t]{l}0.5\end{tabular}}}}%
    \put(0.68215812,0.06495726){\color[rgb]{0.14901961,0.14901961,0.14901961}\makebox(0,0)[lt]{\lineheight{1.25}\smash{\begin{tabular}[t]{l}1\end{tabular}}}}%
    \put(0.94391026,0.06495726){\color[rgb]{0.14901961,0.14901961,0.14901961}\makebox(0,0)[lt]{\lineheight{1.25}\smash{\begin{tabular}[t]{l}1.5\end{tabular}}}}%
    \put(0.32532089,0.02435903){\color[rgb]{0.14901961,0.14901961,0.14901961}\makebox(0,0)[lt]{\lineheight{1.25}\smash{\begin{tabular}[t]{l}Average displacement error in m\end{tabular}}}}%
    \put(0,0){\includegraphics[width=\unitlength,page=2]{images/Nico_Fig_Runtime_over_ADE_v02.pdf}}%
    \put(0.05128205,0.09615385){\color[rgb]{0.14901961,0.14901961,0.14901961}\makebox(0,0)[lt]{\lineheight{1.25}\smash{\begin{tabular}[t]{l}10\end{tabular}}}}%
    \put(0.09615385,0.11538462){\color[rgb]{0.14901961,0.14901961,0.14901961}\makebox(0,0)[lt]{\lineheight{1.25}\smash{\begin{tabular}[t]{l}-1\end{tabular}}}}%
    \put(0.06089744,0.25320513){\color[rgb]{0.14901961,0.14901961,0.14901961}\makebox(0,0)[lt]{\lineheight{1.25}\smash{\begin{tabular}[t]{l}10\end{tabular}}}}%
    \put(0.10576923,0.2724359){\color[rgb]{0.14901961,0.14901961,0.14901961}\makebox(0,0)[lt]{\lineheight{1.25}\smash{\begin{tabular}[t]{l}0\end{tabular}}}}%
    \put(0.06089744,0.41025641){\color[rgb]{0.14901961,0.14901961,0.14901961}\makebox(0,0)[lt]{\lineheight{1.25}\smash{\begin{tabular}[t]{l}10\end{tabular}}}}%
    \put(0.10576923,0.42948718){\color[rgb]{0.14901961,0.14901961,0.14901961}\makebox(0,0)[lt]{\lineheight{1.25}\smash{\begin{tabular}[t]{l}1\end{tabular}}}}%
    \put(0.06089744,0.56730769){\color[rgb]{0.14901961,0.14901961,0.14901961}\makebox(0,0)[lt]{\lineheight{1.25}\smash{\begin{tabular}[t]{l}10\end{tabular}}}}%
    \put(0.10576923,0.58653846){\color[rgb]{0.14901961,0.14901961,0.14901961}\makebox(0,0)[lt]{\lineheight{1.25}\smash{\begin{tabular}[t]{l}2\end{tabular}}}}%
    \put(0.06089744,0.72435897){\color[rgb]{0.14901961,0.14901961,0.14901961}\makebox(0,0)[lt]{\lineheight{1.25}\smash{\begin{tabular}[t]{l}10\end{tabular}}}}%
    \put(0.10576923,0.74358974){\color[rgb]{0.14901961,0.14901961,0.14901961}\makebox(0,0)[lt]{\lineheight{1.25}\smash{\begin{tabular}[t]{l}3\end{tabular}}}}%
    \put(0.04038462,0.34134647){\color[rgb]{0.14901961,0.14901961,0.14901961}\rotatebox{90.00000248}{\makebox(0,0)[lt]{\lineheight{1.25}\smash{\begin{tabular}[t]{l}Runtime in ms\end{tabular}}}}}%
    \put(0,0){\includegraphics[width=\unitlength,page=3]{images/Nico_Fig_Runtime_over_ADE_v02.pdf}}%
    \put(0.51679487,0.54750791){\color[rgb]{0,0,0}\makebox(0,0)[lt]{\lineheight{1.25}\smash{\begin{tabular}[t]{l}AgentF.\end{tabular}}}}%
    \put(0.35752131,0.15101335){\color[rgb]{0,0,0}\makebox(0,0)[lt]{\lineheight{1.25}\smash{\begin{tabular}[t]{l}CVM\end{tabular}}}}%
    \put(0.38448713,0.37376181){\color[rgb]{0,0,0}\makebox(0,0)[lt]{\lineheight{1.25}\smash{\begin{tabular}[t]{l}Social-GAN\end{tabular}}}}%
    \put(0.36491453,0.31790251){\color[rgb]{0,0,0}\makebox(0,0)[lt]{\lineheight{1.25}\smash{\begin{tabular}[t]{l}Social-Impl.\end{tabular}}}}%
    \put(0.25961538,0.61795353){\color[rgb]{0,0,0}\makebox(0,0)[lt]{\lineheight{1.25}\smash{\begin{tabular}[t]{l}Trajectron++\end{tabular}}}}%
    \put(0.65837607,0.48862532){\color[rgb]{0,0,0}\makebox(0,0)[lt]{\lineheight{1.25}\smash{\begin{tabular}[t]{l}Y-Net\end{tabular}}}}%
    \put(0,0){\includegraphics[width=\unitlength,page=4]{images/Nico_Fig_Runtime_over_ADE_v02.pdf}}%
    \put(0.82051282,0.68187788){\color[rgb]{0,0,0}\makebox(0,0)[lt]{\lineheight{1.25}\smash{\begin{tabular}[t]{l}IQR\end{tabular}}}}%
    \put(0,0){\includegraphics[width=\unitlength,page=5]{images/Nico_Fig_Runtime_over_ADE_v02.pdf}}%
    \put(0.82051282,0.63222468){\color[rgb]{0,0,0}\makebox(0,0)[lt]{\lineheight{1.25}\smash{\begin{tabular}[t]{l}Median\end{tabular}}}}%
    \put(0,0){\includegraphics[width=\unitlength,page=6]{images/Nico_Fig_Runtime_over_ADE_v02.pdf}}%
  \end{picture}%
\endgroup%

%% file: chapters/05_discussion.tex
\section{Discussion}
\label{sec_disc}

\subsection{Quantitative analysis}
\label{subsec:quant_anal}
Based on the results gathered in the previous chapter, the strengths and weaknesses of investigated approaches are revealed when analyzed for practical applicability. In terms of accuracy, AgentFormer and Trajectron++ perform best in the Best-of-N evaluation, indicating their capability to generate plausible and multimodal trajectories across diverse scenarios. For the most part, this might be attributed to the Conditional Variational Autoencoder (CVAE) employed in both architectures, as this has been shown to result in more diverse behavior \cite{sohn_learning_2015}. Although both models additionally incorporate an attention-based mechanism to encode relevant interactions in the scene, the mechanism implemented in AgentFormer had minimal to no impact on the overall accuracy when evaluated in this study. When considering only the most likely trajectory, the accuracy of AgentFormer decreases drastically while Trajectron++ still manages to score first among the investigated approaches. This behavior can be partially attributed to the different evaluation procedures included in the latter, returning the most likely path from the employed Gaussian Mixture Model. Following Trajectron++, Social Implicit scores second without this additional evaluation step. Notably, both of these methods employ graph-based networks, highlighting the potential advantages of utilizing this architecture to consider spatial interactions. The temporal modeling on the other hand doesn't seem to have a significant impact as both models utilize a maximum of two timesteps. Hence, the recurrent architecture as well as the 1D temporal graph convolution don't unfold their full potential, leaving room for improvement. Based on the results obtained from the CVM it seems disproportionate to incorporate the entire motion history in the design of future models, but we argue that the consideration is still relevant as will be shown in \cref{subsec:qual_anal}. Since Y-Net and AgentFormer are the only models not explicitly processing relative displacements or velocities, their differing behavior can be attributed to the missing orientational information when provided with a single timestep. To make reasonable predictions, at least two timesteps should be provided here, resulting in performance more similar to the single-timestep case for the remaining models. Therefore, architectures that incorporate object velocities alongside positional information can be seen as beneficial for practical applications, but it remains a challenge to estimate these precisely based just on raw sensor data. 

In terms of execution time, we have analyzed the average inference of the models and evaluated their scalability by applying them to scenarios with varying pedestrian counts. While Trajectron++ has demonstrated the best accuracy, it lags behind in terms of inference time primarily due to its reliance on an LSTM-based architecture. On the contrary, feed-forward neural networks (FNN) hold the promise of significantly faster execution times. This is precisely why Social-Implicit stands out as the network with the fastest inference speed among the investigated models. In addition to this architectural choice, its graph representation ensures robust scalability, making it naturally adaptable to varying agent counts. Nevertheless, while Y-Net employs CNNs within its Encoder-Decoder architecture as well, it exhibits considerably longer runtimes and a less favorable scaling behavior as indicated in appendix \ref{app:eval_runtime}. While the slow inference is caused by the model size, counting over 50 million parameters in total, its spreads can be attributed to the prediction of a single pedestrian at a time, leading to a linear increase in runtime. Similarly, the transformer architecture employed in AgentFormer counts over 6 million parameters, resulting in an equally high runtime but offering better scalability due to the generation of multiple trajectories at a time. Despite SGAN using a recurrent architecture based on LSTM cells, it still shows a fast inference across all scenarios due to its small size, highlighting the interdependency between model size, output generation, and architecture once again. Overall, while a parallel generation of trajectories is more suitable as it requires only a single inference pass and therefore leads to faster runtimes, a sequential approach provides flexibility when adapting to changes in the prediction horizon.

Nevertheless, when considering the trade-off between accuracy and runtime visualized in Fig.~\ref{fig:runtime}, the CVM emerges as the most suitable method on the dataset used. This is in contrast to the current development outlined in chapter \ref{sec_rel_work} where Y-Net, Trajectron++, and AgentFormer were considered state-of-the-art models based on the Best-of-N evaluation approach. As these results suggest that simplistic models, such as the CVM, continue to remain competitive compared to more sophisticated, learning-based approaches, the next chapter will focus on a qualitative analysis, trying to explain this discrepancy.

\subsection{Qualitative analysis}
\label{subsec:qual_anal}

\begin{figure}[b!]
    \begin{center}
    \small{a) Linear prediction}
    \begin{tabular}{c}
        \centering
        \fontsize{8pt}{12pt}\selectfont
        \def\svgwidth{0.9\linewidth}
	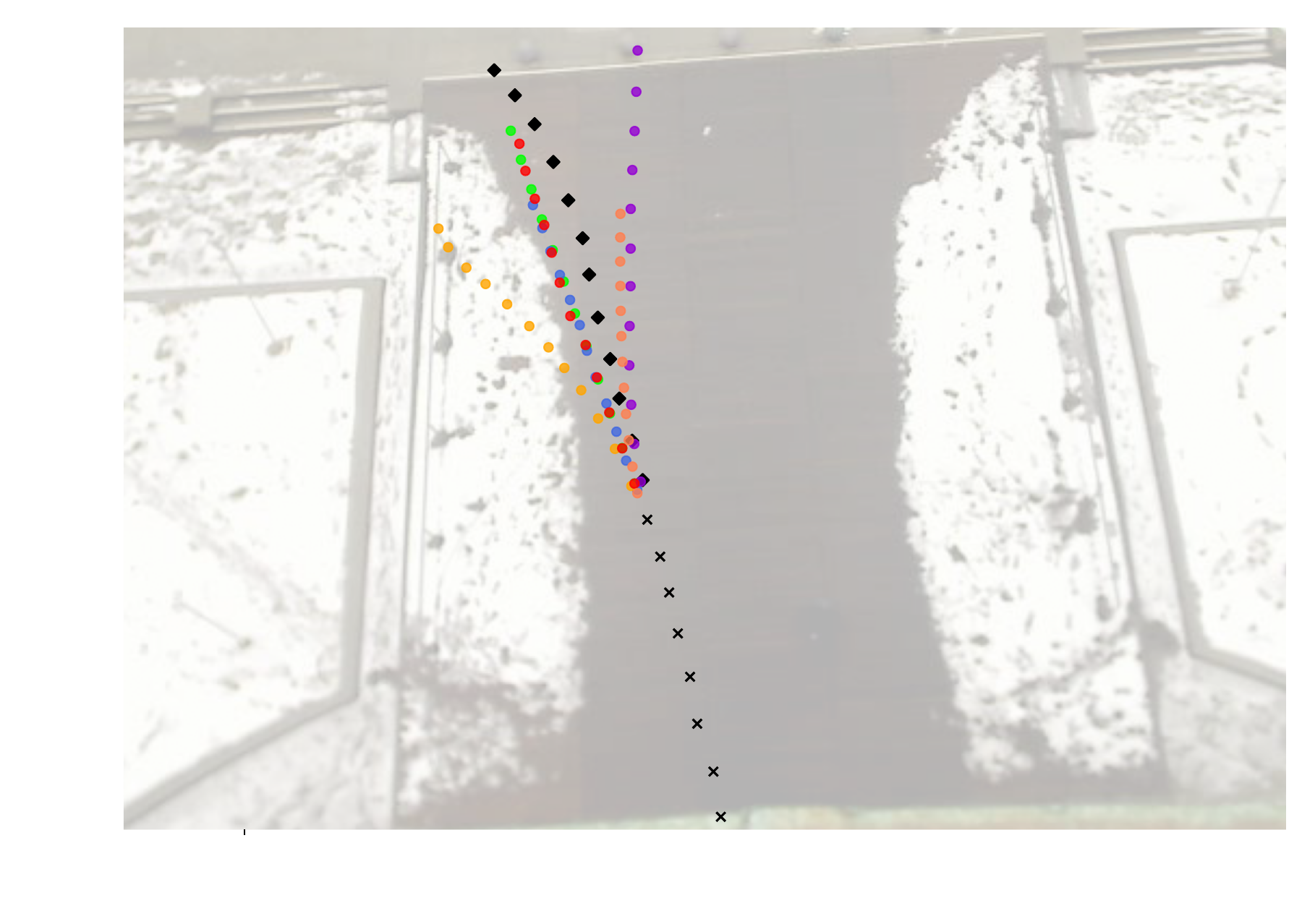
    \end{tabular}
    \end{center}
    
    \begin{center}
    \small{b) Non-linear prediction}
    \begin{tabular}{c}
        \centering
        \fontsize{8pt}{12pt}\selectfont
        \def\svgwidth{0.9\linewidth}
        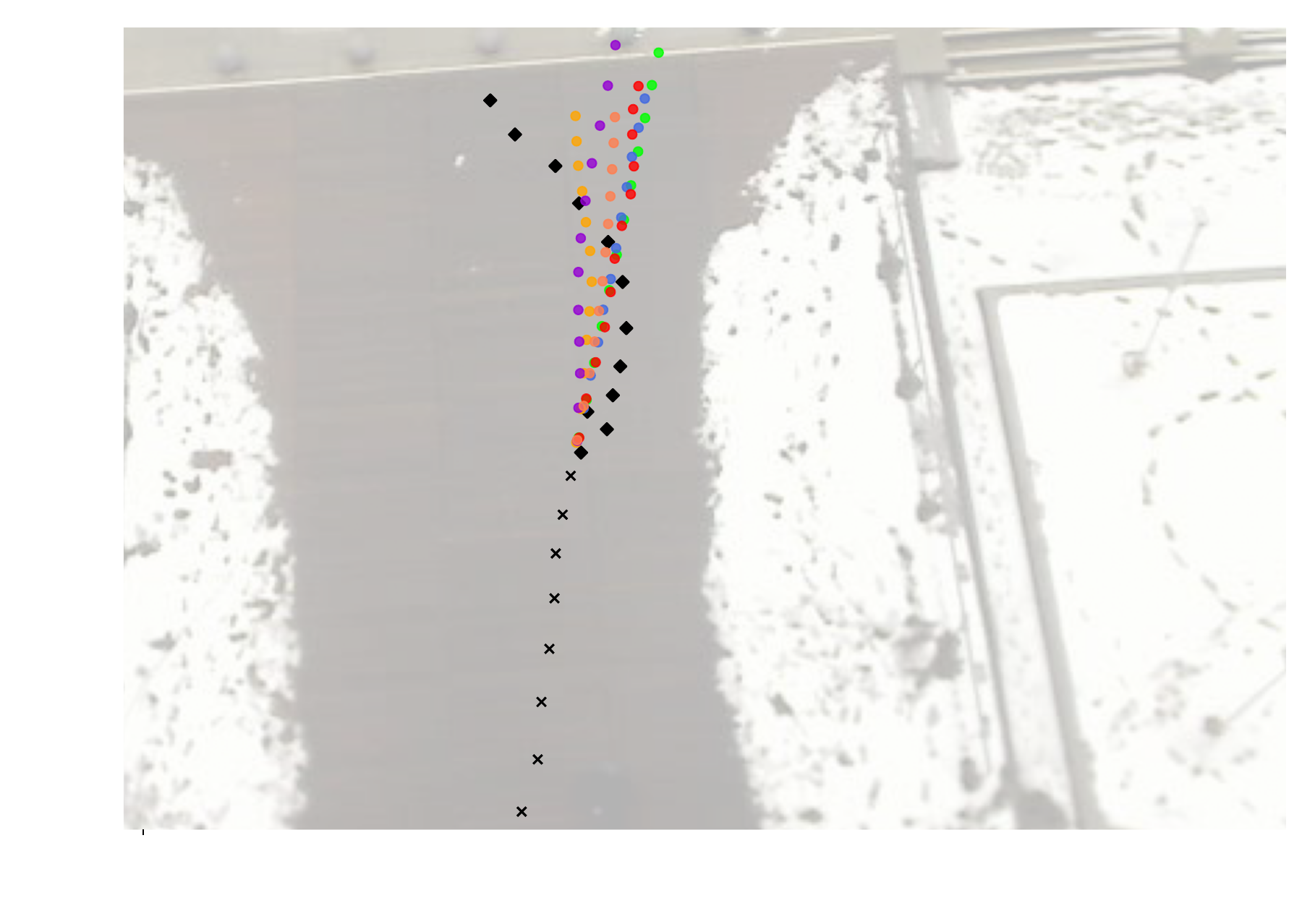
    \end{tabular}
    \end{center}
\end{figure}

\begin{figure}[t!]

    \begin{center}
    \small{c) Static prediction}
    \begin{tabular}{c}
        \centering
        \fontsize{8pt}{12pt}\selectfont
        \def\svgwidth{0.9\linewidth}
	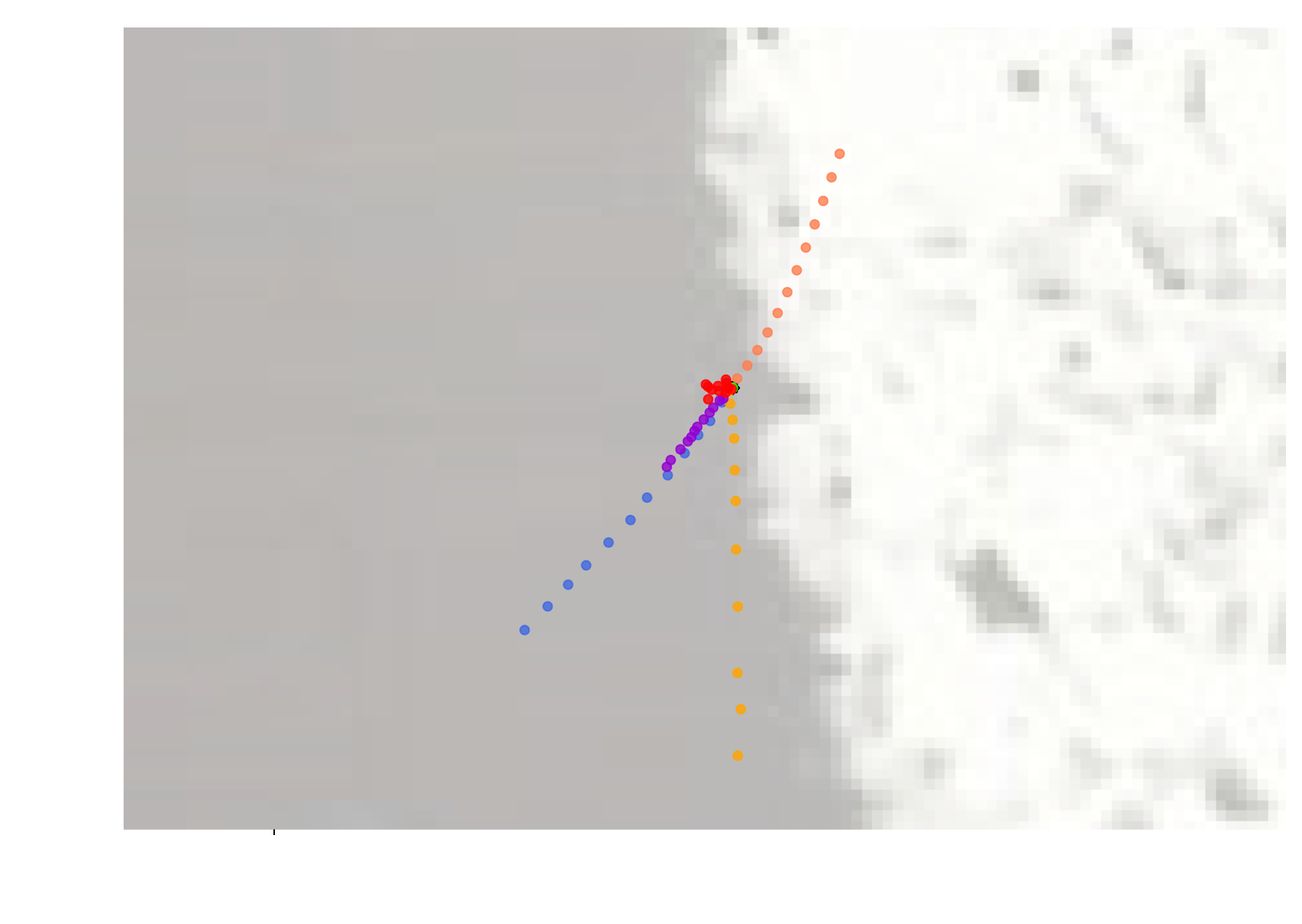
    \end{tabular}
    \end{center}
    
    \begin{center}
    \small{d) State-change prediction}
    \begin{tabular}{c}
        \centering
        \fontsize{8pt}{12pt}\selectfont
        \def\svgwidth{0.9\linewidth}
        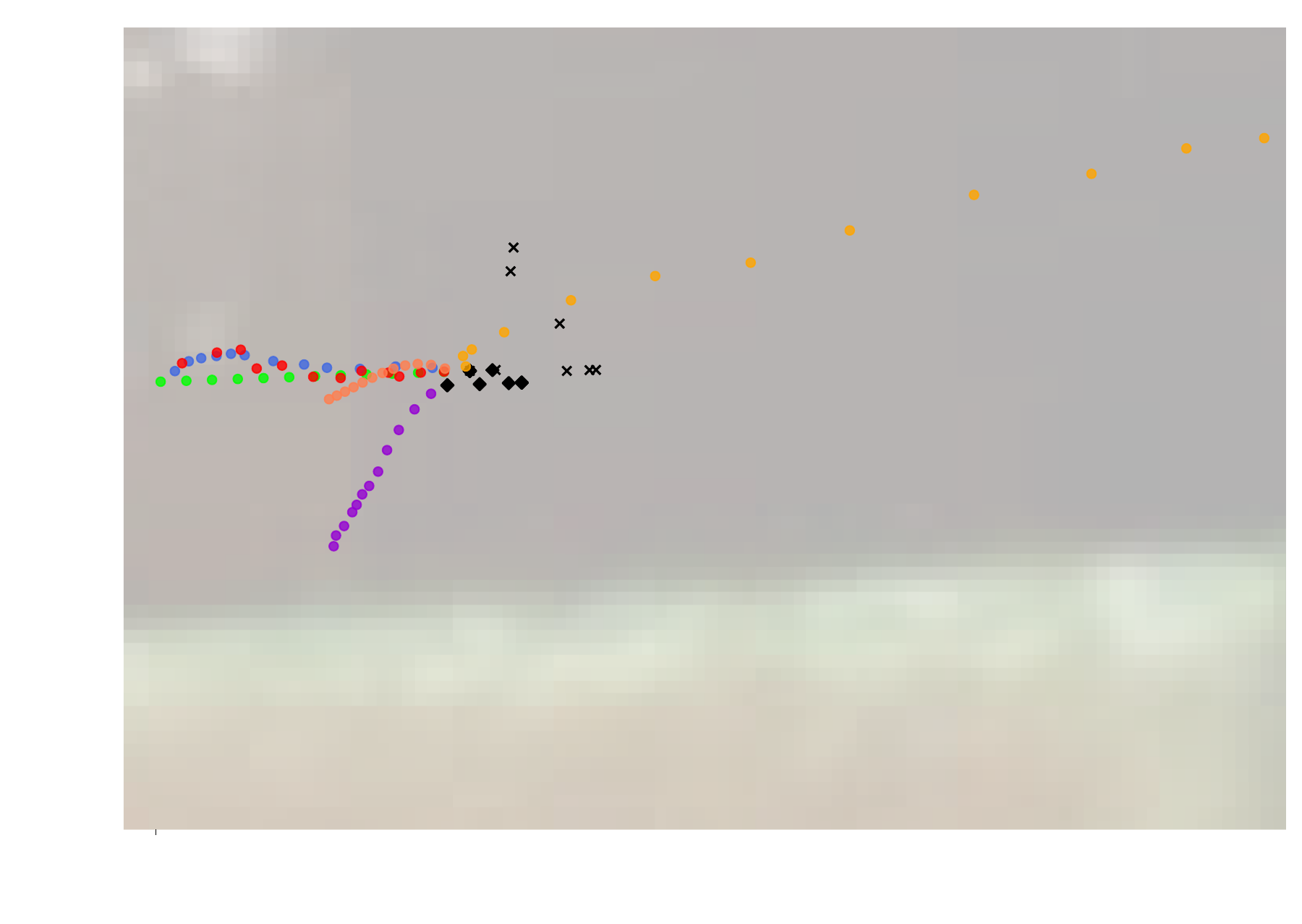
    \end{tabular}
    \end{center}
    
    \caption{Selected samples from the ETH scenario, showcasing four commonly observed pedestrian motion patterns alongside the predictions of each investigated model.}
    \label{fig:trajectory}
\end{figure}

For the analysis provided, representative samples of the ETH/UCY dataset were selected and categorized into four commonly observed prediction scenarios highlighted in figure~\ref{fig:trajectory}. The first category visualized with figure~\ref{fig:trajectory}a captures situations where pedestrians move in an overall linear manner within the prediction horizon. For these scenes, a directional spread for the individual approaches is noticeable, whereas the constant velocity model and Trajectron++ manage to generate the most accurate predictions. In general, the simple CVM already provides a good approximation for these cases, but they only account for a small fraction of scenarios within the dataset \cite{kothari_human_2022}. Hence, non-linear predictions represent the majority of cases with the second category where a challenging scene is visualized in figure~\ref{fig:trajectory}b. Although neural networks offer a distinct advantage when generating non-linear trajectories \cite{mohamed_social-implicit_2022}, none of the investigated models fully capture the dynamics shown. While these situations are considered unpredictable due to the inherent epistemic uncertainty involved \cite{mangalam_goals_2020}, all models manage to capture the overall movement direction of the scene. Along with the first category, these types of errors occur on an operational level, where the overall intention is detected and only minor variations remain \cite{vizzari_agent_2020}. To improve the prediction in these situations, the inner state of the agent would need to be considered, which is difficult to determine by any outside observer \cite{rasouli_autonomous_2018}. Moreover, drawing from the field of evacuation dynamics, cultural factors, among many more, might play an important role as well \cite{schadschneider_evacuation_2009}. While within this dataset, a linear prediction often provides a reasonable approximation for non-linear cases, neural networks have a distinct advantage in situations where environmental factors have a strong influence on decision-making \cite{gorrini_observation_2018}.

As the categories a) and b) solely focus on dynamic agents, the third one comprises static ones as displayed in figure~\ref{fig:trajectory}c. These are commonly encountered in urban scenarios and are the most prominent cause for the performance differences observed. Notably, all approaches analyzed tend to generate trajectories even in the absence of any positional changes between successive timesteps. Furthermore, the absence of any directional information results in the prediction of linear trajectories with various orientations. This kind of error is different from the one observed in the first two categories as it reflects the inability of some methods to reason about a pedestrian's goal or intent. Since human behavior is inherently goal-oriented, the resulting trajectory is strongly influenced by this aspect \cite{vizzari_agent_2020, mangalam_goals_2020}. On a data level, this bias can be explained by the under-representation of static scenarios within the given dataset. Despite both AgentFormer and Y-Net considering additional timesteps in contrast to the other approaches, as well we Y-Net reasoning about one's goal, they do not appear to effectively leverage this information. In the sample presented, the CVM provides the most precise estimation, followed by Trajectron++, which might hint at its capability to reason about different velocity profiles. In comparison, Social-Implicit is designed to explicitly handle pedestrians with varying velocity profiles, consisting of an ensemble architecture trained on subsets of the original dataset for different speeds \cite{mohamed_social-implicit_2022}. Nevertheless, although this architecture models this motion-less group, it still exhibits this bias. \\

The fourth and last prediction category is represented by state-changes where an agent changes its motion state between the observation and the ground truth as highlighted in image~\ref{fig:trajectory}d. These situations are often found in traffic environments and approaches intended to be used in autonomous systems should be capable of handling those. Here, all investigated approaches fail to identify the intention of the pedestrians in the presented case. Although such behavior might be predictable when taking into account spatial information in urban environments, it still poses a challenge even for human drivers \cite{rasouli_pie_2019}. To the best of our knowledge, these situations have not been explicitly addressed by trajectory prediction research and highlight the significance of considering additional information like motion history and semantic clues besides just the last timestep \cite{rasouli_autonomous_2018}. While for most dynamic and static cases, the constant velocity model provides a reasonable approximation, it reaches its limits for the non-homogeneous movement represented by this category as well as real-world urban situations related to this \cite{gorrini_observation_2018}.

\subsection{Limitations}
Given the results presented, the question can be raised whether pedestrians predominantly move linearly and thus for the dynamic cases, the CVM delivers a good approximation of the average walking behavior observed. This hypothesis seems to be supported by the results showcased for Trajectron++. Nevertheless, it remains uncertain whether the findings presented in this study can be generalized to other datasets such as SDD \cite{robicquet_learning_2016} or nuScenes \cite{caesar_nuscenes_2020} as contextual information plays an important role in decision-making \cite{gorrini_observation_2018}. Hence, it is to be expected that when evaluating more complex traffic environments, the CVM will perform significantly worse for non-linear cases. As the ETH/UCY dataset exclusively comprises pedestrian data, this paper presents a general evaluation of the ability of certain methods to generate single pedestrian trajectories when confronted with the requirements of autonomous systems. To apply trajectory prediction algorithms to real-world applications, the consideration of other road users as well as the road infrastructure is vital and requires a deeper investigation \cite{rasouli_autonomous_2018}. Regardless, the consideration of non-homogeneous movements represents a significant challenge and an open research topic. In addition, it needs to be highlighted that with the employed Leave-One-Out Cross-Validation, five models were individually trained and tested, whereas in practical applications only one model should be utilized. Furthermore, the transferability of prediction methods across different scenes and datasets remains a challenge and an active area of research \cite{ivanovic_expanding_2023}. 

\subsection{Future research}
As represented through the quantitative and qualitative analysis outlined in this chapter, open challenges remain that need to be addressed before applying these systems in an autonomous vehicle. While the CVM seems to provide a good tool for approximating the most likely trajectory for dynamic pedestrians, it shows weaknesses for state-changes and non-homogeneous cases. Such situations frequently occur in urban scenarios and require a deeper scene understanding which can only be reflected through learning-based approaches. Consequently, the focus should be shifted to the development of approaches that fulfill two key criteria: First, the integration of spatial information alongside an agent's motion history, and second, the effective utilization of these features, as not all architectures currently achieve this. These considerations also play an important role in improving the overall reliability of these methods as the detections used are usually derived from noisy sensor data. In addition, the focus should be shifted towards traffic-oriented datasets to better align with situations encountered in real-world applications. As for the model architecture, the development of hybrid approaches can also be considered, training a network explicitly for the detection of state changes and static cases, while using the CVM to approximate the overall dynamics. 

%% file: images/4.pdf_tex
\begingroup%
  \makeatletter%
  \providecommand\color[2][]{%
    \errmessage{(Inkscape) Color is used for the text in Inkscape, but the package 'color.sty' is not loaded}%
    \renewcommand\color[2][]{}%
  }%
  \providecommand\transparent[1]{%
    \errmessage{(Inkscape) Transparency is used (non-zero) for the text in Inkscape, but the package 'transparent.sty' is not loaded}%
    \renewcommand\transparent[1]{}%
  }%
  \providecommand\rotatebox[2]{#2}%
  \newcommand*\fsize{\dimexpr\f@size pt\relax}%
  \newcommand*\lineheight[1]{\fontsize{\fsize}{#1\fsize}\selectfont}%
  \ifx\svgwidth\undefined%
    \setlength{\unitlength}{863.9999654bp}%
    \ifx\svgscale\undefined%
      \relax%
    \else%
      \setlength{\unitlength}{\unitlength * \real{\svgscale}}%
    \fi%
  \else%
    \setlength{\unitlength}{\svgwidth}%
  \fi%
  \global\let\svgwidth\undefined%
  \global\let\svgscale\undefined%
  \makeatother%
  \begin{picture}(1,0.70833334)%
    \lineheight{1}%
    \setlength\tabcolsep{0pt}%
    \put(0,0){\includegraphics[width=\unitlength,page=1]{images/4.pdf}}%
    \put(0.16469724,0.04533201){\makebox(0,0)[lt]{\lineheight{1.25}\smash{\begin{tabular}[t]{l}100\end{tabular}}}}%
    \put(0,0){\includegraphics[width=\unitlength,page=2]{images/4.pdf}}%
    \put(0.33057288,0.04533201){\makebox(0,0)[lt]{\lineheight{1.25}\smash{\begin{tabular}[t]{l}200\end{tabular}}}}%
    \put(0,0){\includegraphics[width=\unitlength,page=3]{images/4.pdf}}%
    \put(0.49644853,0.04533201){\makebox(0,0)[lt]{\lineheight{1.25}\smash{\begin{tabular}[t]{l}300\end{tabular}}}}%
    \put(0,0){\includegraphics[width=\unitlength,page=4]{images/4.pdf}}%
    \put(0.66232418,0.04533201){\makebox(0,0)[lt]{\lineheight{1.25}\smash{\begin{tabular}[t]{l}400\end{tabular}}}}%
    \put(0,0){\includegraphics[width=\unitlength,page=5]{images/4.pdf}}%
    \put(0.82819982,0.04533201){\makebox(0,0)[lt]{\lineheight{1.25}\smash{\begin{tabular}[t]{l}500\end{tabular}}}}%
    \put(0.51301109,0.01717446){\makebox(0,0)[lt]{\lineheight{1.25}\smash{\begin{tabular}[t]{l}Pixel\end{tabular}}}}%
    \put(0,0){\includegraphics[width=\unitlength,page=6]{images/4.pdf}}%
    \put(0.04022209,0.65775994){\makebox(0,0)[lt]{\lineheight{1.25}\smash{\begin{tabular}[t]{l}100\end{tabular}}}}%
    \put(0,0){\includegraphics[width=\unitlength,page=7]{images/4.pdf}}%
    \put(0.04022209,0.57482211){\makebox(0,0)[lt]{\lineheight{1.25}\smash{\begin{tabular}[t]{l}150\end{tabular}}}}%
    \put(0,0){\includegraphics[width=\unitlength,page=8]{images/4.pdf}}%
    \put(0.04022209,0.49188429){\makebox(0,0)[lt]{\lineheight{1.25}\smash{\begin{tabular}[t]{l}200\end{tabular}}}}%
    \put(0,0){\includegraphics[width=\unitlength,page=9]{images/4.pdf}}%
    \put(0.04022209,0.40894646){\makebox(0,0)[lt]{\lineheight{1.25}\smash{\begin{tabular}[t]{l}250\end{tabular}}}}%
    \put(0,0){\includegraphics[width=\unitlength,page=10]{images/4.pdf}}%
    \put(0.04022209,0.32600864){\makebox(0,0)[lt]{\lineheight{1.25}\smash{\begin{tabular}[t]{l}300\end{tabular}}}}%
    \put(0,0){\includegraphics[width=\unitlength,page=11]{images/4.pdf}}%
    \put(0.04022209,0.24307081){\makebox(0,0)[lt]{\lineheight{1.25}\smash{\begin{tabular}[t]{l}350\end{tabular}}}}%
    \put(0,0){\includegraphics[width=\unitlength,page=12]{images/4.pdf}}%
    \put(0.04022209,0.16013299){\makebox(0,0)[lt]{\lineheight{1.25}\smash{\begin{tabular}[t]{l}400\end{tabular}}}}%
    \put(0,0){\includegraphics[width=\unitlength,page=13]{images/4.pdf}}%
    \put(0.04022209,0.07719516){\makebox(0,0)[lt]{\lineheight{1.25}\smash{\begin{tabular}[t]{l}450\end{tabular}}}}%
    \put(0.0305288,0.35147222){\rotatebox{90}{\makebox(0,0)[lt]{\lineheight{1.25}\smash{\begin{tabular}[t]{l}Pixel\end{tabular}}}}}%
    \put(0,0){\includegraphics[width=\unitlength,page=14]{images/4.pdf}}%
    \put(0.78,0.64496776){\makebox(0,0)[lt]{\lineheight{1.25}\smash{\begin{tabular}[t]{l}Ground truth\end{tabular}}}}%
    \put(0,0){\includegraphics[width=\unitlength,page=15]{images/4.pdf}}%
    \put(0.78,0.60758711){\makebox(0,0)[lt]{\lineheight{1.25}\smash{\begin{tabular}[t]{l}Observation\end{tabular}}}}%
    \put(0,0){\includegraphics[width=\unitlength,page=16]{images/4.pdf}}%
    \put(0.78,0.57020647){\makebox(0,0)[lt]{\lineheight{1.25}\smash{\begin{tabular}[t]{l}CVM\end{tabular}}}}%
    \put(0,0){\includegraphics[width=\unitlength,page=17]{images/4.pdf}}%
    \put(0.78,0.53282582){\makebox(0,0)[lt]{\lineheight{1.25}\smash{\begin{tabular}[t]{l}Social-Impl.\end{tabular}}}}%
    \put(0,0){\includegraphics[width=\unitlength,page=18]{images/4.pdf}}%
    \put(0.78,0.49544518){\makebox(0,0)[lt]{\lineheight{1.25}\smash{\begin{tabular}[t]{l}AgentFormer\end{tabular}}}}%
    \put(0,0){\includegraphics[width=\unitlength,page=19]{images/4.pdf}}%
    \put(0.78,0.45806453){\makebox(0,0)[lt]{\lineheight{1.25}\smash{\begin{tabular}[t]{l}Ynet\end{tabular}}}}%
    \put(0,0){\includegraphics[width=\unitlength,page=20]{images/4.pdf}}%
    \put(0.78,0.42068389){\makebox(0,0)[lt]{\lineheight{1.25}\smash{\begin{tabular}[t]{l}Traj++\end{tabular}}}}%
    \put(0,0){\includegraphics[width=\unitlength,page=21]{images/4.pdf}}%
    \put(0.78,0.38330324){\makebox(0,0)[lt]{\lineheight{1.25}\smash{\begin{tabular}[t]{l}Sgan\end{tabular}}}}%
  \end{picture}%
\endgroup%

%% file: images/1.pdf_tex
\begingroup%
  \makeatletter%
  \providecommand\color[2][]{%
    \errmessage{(Inkscape) Color is used for the text in Inkscape, but the package 'color.sty' is not loaded}%
    \renewcommand\color[2][]{}%
  }%
  \providecommand\transparent[1]{%
    \errmessage{(Inkscape) Transparency is used (non-zero) for the text in Inkscape, but the package 'transparent.sty' is not loaded}%
    \renewcommand\transparent[1]{}%
  }%
  \providecommand\rotatebox[2]{#2}%
  \newcommand*\fsize{\dimexpr\f@size pt\relax}%
  \newcommand*\lineheight[1]{\fontsize{\fsize}{#1\fsize}\selectfont}%
  \ifx\svgwidth\undefined%
    \setlength{\unitlength}{863.9999654bp}%
    \ifx\svgscale\undefined%
      \relax%
    \else%
      \setlength{\unitlength}{\unitlength * \real{\svgscale}}%
    \fi%
  \else%
    \setlength{\unitlength}{\svgwidth}%
  \fi%
  \global\let\svgwidth\undefined%
  \global\let\svgscale\undefined%
  \makeatother%
  \begin{picture}(1,0.70833334)%
    \lineheight{1}%
    \setlength\tabcolsep{0pt}%
    \put(0,0){\includegraphics[width=\unitlength,page=1]{images/1.pdf}}%
    \put(0.08670195,0.04533201){\makebox(0,0)[lt]{\lineheight{1.25}\smash{\begin{tabular}[t]{l}200\end{tabular}}}}%
    \put(0,0){\includegraphics[width=\unitlength,page=2]{images/1.pdf}}%
    \put(0.1933911,0.04533201){\makebox(0,0)[lt]{\lineheight{1.25}\smash{\begin{tabular}[t]{l}250\end{tabular}}}}%
    \put(0,0){\includegraphics[width=\unitlength,page=3]{images/1.pdf}}%
    \put(0.30008025,0.04533201){\makebox(0,0)[lt]{\lineheight{1.25}\smash{\begin{tabular}[t]{l}300\end{tabular}}}}%
    \put(0,0){\includegraphics[width=\unitlength,page=4]{images/1.pdf}}%
    \put(0.4067694,0.04533201){\makebox(0,0)[lt]{\lineheight{1.25}\smash{\begin{tabular}[t]{l}350\end{tabular}}}}%
    \put(0,0){\includegraphics[width=\unitlength,page=5]{images/1.pdf}}%
    \put(0.51345856,0.04533201){\makebox(0,0)[lt]{\lineheight{1.25}\smash{\begin{tabular}[t]{l}400\end{tabular}}}}%
    \put(0,0){\includegraphics[width=\unitlength,page=6]{images/1.pdf}}%
    \put(0.62014772,0.04533201){\makebox(0,0)[lt]{\lineheight{1.25}\smash{\begin{tabular}[t]{l}450\end{tabular}}}}%
    \put(0,0){\includegraphics[width=\unitlength,page=7]{images/1.pdf}}%
    \put(0.72683687,0.04533201){\makebox(0,0)[lt]{\lineheight{1.25}\smash{\begin{tabular}[t]{l}500\end{tabular}}}}%
    \put(0,0){\includegraphics[width=\unitlength,page=8]{images/1.pdf}}%
    \put(0.83352602,0.04533201){\makebox(0,0)[lt]{\lineheight{1.25}\smash{\begin{tabular}[t]{l}550\end{tabular}}}}%
    \put(0,0){\includegraphics[width=\unitlength,page=9]{images/1.pdf}}%
    \put(0.94021514,0.04533201){\makebox(0,0)[lt]{\lineheight{1.25}\smash{\begin{tabular}[t]{l}600\end{tabular}}}}%
    \put(0.51301109,0.01717446){\makebox(0,0)[lt]{\lineheight{1.25}\smash{\begin{tabular}[t]{l}Pixel\end{tabular}}}}%
    \put(0,0){\includegraphics[width=\unitlength,page=10]{images/1.pdf}}%
    \put(0.04022209,0.65204914){\makebox(0,0)[lt]{\lineheight{1.25}\smash{\begin{tabular}[t]{l}100\end{tabular}}}}%
    \put(0,0){\includegraphics[width=\unitlength,page=11]{images/1.pdf}}%
    \put(0.04022209,0.54535998){\makebox(0,0)[lt]{\lineheight{1.25}\smash{\begin{tabular}[t]{l}150\end{tabular}}}}%
    \put(0,0){\includegraphics[width=\unitlength,page=12]{images/1.pdf}}%
    \put(0.04022209,0.43867082){\makebox(0,0)[lt]{\lineheight{1.25}\smash{\begin{tabular}[t]{l}200\end{tabular}}}}%
    \put(0,0){\includegraphics[width=\unitlength,page=13]{images/1.pdf}}%
    \put(0.04022209,0.33198167){\makebox(0,0)[lt]{\lineheight{1.25}\smash{\begin{tabular}[t]{l}250\end{tabular}}}}%
    \put(0,0){\includegraphics[width=\unitlength,page=14]{images/1.pdf}}%
    \put(0.04022209,0.22529251){\makebox(0,0)[lt]{\lineheight{1.25}\smash{\begin{tabular}[t]{l}300\end{tabular}}}}%
    \put(0,0){\includegraphics[width=\unitlength,page=15]{images/1.pdf}}%
    \put(0.04022209,0.11860336){\makebox(0,0)[lt]{\lineheight{1.25}\smash{\begin{tabular}[t]{l}350\end{tabular}}}}%
    \put(0.0305288,0.35147222){\rotatebox{90}{\makebox(0,0)[lt]{\lineheight{1.25}\smash{\begin{tabular}[t]{l}Pixel\end{tabular}}}}}%
    \put(0,0){\includegraphics[width=\unitlength,page=16]{images/1.pdf}}%
    \put(0.78,0.64496776){\makebox(0,0)[lt]{\lineheight{1.25}\smash{\begin{tabular}[t]{l}Ground truth\end{tabular}}}}%
    \put(0,0){\includegraphics[width=\unitlength,page=17]{images/1.pdf}}%
    \put(0.78,0.60758711){\makebox(0,0)[lt]{\lineheight{1.25}\smash{\begin{tabular}[t]{l}Observation\end{tabular}}}}%
    \put(0,0){\includegraphics[width=\unitlength,page=18]{images/1.pdf}}%
    \put(0.78,0.57020647){\makebox(0,0)[lt]{\lineheight{1.25}\smash{\begin{tabular}[t]{l}CVM\end{tabular}}}}%
    \put(0,0){\includegraphics[width=\unitlength,page=19]{images/1.pdf}}%
    \put(0.78,0.53282582){\makebox(0,0)[lt]{\lineheight{1.25}\smash{\begin{tabular}[t]{l}Social-Impl.\end{tabular}}}}%
    \put(0,0){\includegraphics[width=\unitlength,page=20]{images/1.pdf}}%
    \put(0.78,0.49544518){\makebox(0,0)[lt]{\lineheight{1.25}\smash{\begin{tabular}[t]{l}AgentFormer\end{tabular}}}}%
    \put(0,0){\includegraphics[width=\unitlength,page=21]{images/1.pdf}}%
    \put(0.78,0.45806453){\makebox(0,0)[lt]{\lineheight{1.25}\smash{\begin{tabular}[t]{l}Ynet\end{tabular}}}}%
    \put(0,0){\includegraphics[width=\unitlength,page=22]{images/1.pdf}}%
    \put(0.78,0.42068389){\makebox(0,0)[lt]{\lineheight{1.25}\smash{\begin{tabular}[t]{l}Traj++\end{tabular}}}}%
    \put(0,0){\includegraphics[width=\unitlength,page=23]{images/1.pdf}}%
    \put(0.78,0.38330324){\makebox(0,0)[lt]{\lineheight{1.25}\smash{\begin{tabular}[t]{l}Sgan\end{tabular}}}}%
  \end{picture}%
\endgroup%

%% file: images/2.pdf_tex
\begingroup%
  \makeatletter%
  \providecommand\color[2][]{%
    \errmessage{(Inkscape) Color is used for the text in Inkscape, but the package 'color.sty' is not loaded}%
    \renewcommand\color[2][]{}%
  }%
  \providecommand\transparent[1]{%
    \errmessage{(Inkscape) Transparency is used (non-zero) for the text in Inkscape, but the package 'transparent.sty' is not loaded}%
    \renewcommand\transparent[1]{}%
  }%
  \providecommand\rotatebox[2]{#2}%
  \newcommand*\fsize{\dimexpr\f@size pt\relax}%
  \newcommand*\lineheight[1]{\fontsize{\fsize}{#1\fsize}\selectfont}%
  \ifx\svgwidth\undefined%
    \setlength{\unitlength}{863.9999654bp}%
    \ifx\svgscale\undefined%
      \relax%
    \else%
      \setlength{\unitlength}{\unitlength * \real{\svgscale}}%
    \fi%
  \else%
    \setlength{\unitlength}{\svgwidth}%
  \fi%
  \global\let\svgwidth\undefined%
  \global\let\svgscale\undefined%
  \makeatother%
  \begin{picture}(1,0.70833334)%
    \lineheight{1}%
    \setlength\tabcolsep{0pt}%
    \put(0,0){\includegraphics[width=\unitlength,page=1]{images/2.pdf}}%
    \put(0.18735329,0.04533201){\makebox(0,0)[lt]{\lineheight{1.25}\smash{\begin{tabular}[t]{l}360\end{tabular}}}}%
    \put(0,0){\includegraphics[width=\unitlength,page=2]{images/2.pdf}}%
    \put(0.34702702,0.04533201){\makebox(0,0)[lt]{\lineheight{1.25}\smash{\begin{tabular}[t]{l}380\end{tabular}}}}%
    \put(0,0){\includegraphics[width=\unitlength,page=3]{images/2.pdf}}%
    \put(0.50670076,0.04533201){\makebox(0,0)[lt]{\lineheight{1.25}\smash{\begin{tabular}[t]{l}400\end{tabular}}}}%
    \put(0,0){\includegraphics[width=\unitlength,page=4]{images/2.pdf}}%
    \put(0.6663745,0.04533201){\makebox(0,0)[lt]{\lineheight{1.25}\smash{\begin{tabular}[t]{l}420\end{tabular}}}}%
    \put(0,0){\includegraphics[width=\unitlength,page=5]{images/2.pdf}}%
    \put(0.82604823,0.04533201){\makebox(0,0)[lt]{\lineheight{1.25}\smash{\begin{tabular}[t]{l}440\end{tabular}}}}%
    \put(0.51301109,0.01717446){\makebox(0,0)[lt]{\lineheight{1.25}\smash{\begin{tabular}[t]{l}Pixel\end{tabular}}}}%
    \put(0,0){\includegraphics[width=\unitlength,page=6]{images/2.pdf}}%
    \put(0.04022209,0.65629365){\makebox(0,0)[lt]{\lineheight{1.25}\smash{\begin{tabular}[t]{l}280\end{tabular}}}}%
    \put(0,0){\includegraphics[width=\unitlength,page=7]{images/2.pdf}}%
    \put(0.04022209,0.57645678){\makebox(0,0)[lt]{\lineheight{1.25}\smash{\begin{tabular}[t]{l}290\end{tabular}}}}%
    \put(0,0){\includegraphics[width=\unitlength,page=8]{images/2.pdf}}%
    \put(0.04022209,0.49661992){\makebox(0,0)[lt]{\lineheight{1.25}\smash{\begin{tabular}[t]{l}300\end{tabular}}}}%
    \put(0,0){\includegraphics[width=\unitlength,page=9]{images/2.pdf}}%
    \put(0.04022209,0.41678305){\makebox(0,0)[lt]{\lineheight{1.25}\smash{\begin{tabular}[t]{l}310\end{tabular}}}}%
    \put(0,0){\includegraphics[width=\unitlength,page=10]{images/2.pdf}}%
    \put(0.04022209,0.33694617){\makebox(0,0)[lt]{\lineheight{1.25}\smash{\begin{tabular}[t]{l}320\end{tabular}}}}%
    \put(0,0){\includegraphics[width=\unitlength,page=11]{images/2.pdf}}%
    \put(0.04022209,0.2571093){\makebox(0,0)[lt]{\lineheight{1.25}\smash{\begin{tabular}[t]{l}330\end{tabular}}}}%
    \put(0,0){\includegraphics[width=\unitlength,page=12]{images/2.pdf}}%
    \put(0.04022209,0.17727243){\makebox(0,0)[lt]{\lineheight{1.25}\smash{\begin{tabular}[t]{l}340\end{tabular}}}}%
    \put(0,0){\includegraphics[width=\unitlength,page=13]{images/2.pdf}}%
    \put(0.04022209,0.09743556){\makebox(0,0)[lt]{\lineheight{1.25}\smash{\begin{tabular}[t]{l}350\end{tabular}}}}%
    \put(0.0305288,0.35147222){\rotatebox{90}{\makebox(0,0)[lt]{\lineheight{1.25}\smash{\begin{tabular}[t]{l}Pixel\end{tabular}}}}}%
    \put(0,0){\includegraphics[width=\unitlength,page=14]{images/2.pdf}}%
    \put(0.78,0.64496776){\makebox(0,0)[lt]{\lineheight{1.25}\smash{\begin{tabular}[t]{l}Ground truth\end{tabular}}}}%
    \put(0,0){\includegraphics[width=\unitlength,page=15]{images/2.pdf}}%
    \put(0.78,0.60758711){\makebox(0,0)[lt]{\lineheight{1.25}\smash{\begin{tabular}[t]{l}Observation\end{tabular}}}}%
    \put(0,0){\includegraphics[width=\unitlength,page=16]{images/2.pdf}}%
    \put(0.78,0.57020647){\makebox(0,0)[lt]{\lineheight{1.25}\smash{\begin{tabular}[t]{l}CVM\end{tabular}}}}%
    \put(0,0){\includegraphics[width=\unitlength,page=17]{images/2.pdf}}%
    \put(0.78,0.53282582){\makebox(0,0)[lt]{\lineheight{1.25}\smash{\begin{tabular}[t]{l}Social-Impl.\end{tabular}}}}%
    \put(0,0){\includegraphics[width=\unitlength,page=18]{images/2.pdf}}%
    \put(0.78,0.49544518){\makebox(0,0)[lt]{\lineheight{1.25}\smash{\begin{tabular}[t]{l}AgentFormer\end{tabular}}}}%
    \put(0,0){\includegraphics[width=\unitlength,page=19]{images/2.pdf}}%
    \put(0.78,0.45806453){\makebox(0,0)[lt]{\lineheight{1.25}\smash{\begin{tabular}[t]{l}Ynet\end{tabular}}}}%
    \put(0,0){\includegraphics[width=\unitlength,page=20]{images/2.pdf}}%
    \put(0.78,0.42068389){\makebox(0,0)[lt]{\lineheight{1.25}\smash{\begin{tabular}[t]{l}Traj++\end{tabular}}}}%
    \put(0,0){\includegraphics[width=\unitlength,page=21]{images/2.pdf}}%
    \put(0.78,0.38330324){\makebox(0,0)[lt]{\lineheight{1.25}\smash{\begin{tabular}[t]{l}Sgan\end{tabular}}}}%
  \end{picture}%
\endgroup%

%% file: images/3.pdf_tex
\begingroup%
  \makeatletter%
  \providecommand\color[2][]{%
    \errmessage{(Inkscape) Color is used for the text in Inkscape, but the package 'color.sty' is not loaded}%
    \renewcommand\color[2][]{}%
  }%
  \providecommand\transparent[1]{%
    \errmessage{(Inkscape) Transparency is used (non-zero) for the text in Inkscape, but the package 'transparent.sty' is not loaded}%
    \renewcommand\transparent[1]{}%
  }%
  \providecommand\rotatebox[2]{#2}%
  \newcommand*\fsize{\dimexpr\f@size pt\relax}%
  \newcommand*\lineheight[1]{\fontsize{\fsize}{#1\fsize}\selectfont}%
  \ifx\svgwidth\undefined%
    \setlength{\unitlength}{863.9999654bp}%
    \ifx\svgscale\undefined%
      \relax%
    \else%
      \setlength{\unitlength}{\unitlength * \real{\svgscale}}%
    \fi%
  \else%
    \setlength{\unitlength}{\svgwidth}%
  \fi%
  \global\let\svgwidth\undefined%
  \global\let\svgscale\undefined%
  \makeatother%
  \begin{picture}(1,0.70833334)%
    \lineheight{1}%
    \setlength\tabcolsep{0pt}%
    \put(0,0){\includegraphics[width=\unitlength,page=1]{images/3.pdf}}%
    \put(0.09656133,0.04533201){\makebox(0,0)[lt]{\lineheight{1.25}\smash{\begin{tabular}[t]{l}240\end{tabular}}}}%
    \put(0,0){\includegraphics[width=\unitlength,page=2]{images/3.pdf}}%
    \put(0.29036446,0.04533201){\makebox(0,0)[lt]{\lineheight{1.25}\smash{\begin{tabular}[t]{l}260\end{tabular}}}}%
    \put(0,0){\includegraphics[width=\unitlength,page=3]{images/3.pdf}}%
    \put(0.4841676,0.04533201){\makebox(0,0)[lt]{\lineheight{1.25}\smash{\begin{tabular}[t]{l}280\end{tabular}}}}%
    \put(0,0){\includegraphics[width=\unitlength,page=4]{images/3.pdf}}%
    \put(0.67797073,0.04533201){\makebox(0,0)[lt]{\lineheight{1.25}\smash{\begin{tabular}[t]{l}300\end{tabular}}}}%
    \put(0,0){\includegraphics[width=\unitlength,page=5]{images/3.pdf}}%
    \put(0.8717739,0.04533201){\makebox(0,0)[lt]{\lineheight{1.25}\smash{\begin{tabular}[t]{l}320\end{tabular}}}}%
    \put(0.51301109,0.01717446){\makebox(0,0)[lt]{\lineheight{1.25}\smash{\begin{tabular}[t]{l}Pixel\end{tabular}}}}%
    \put(0,0){\includegraphics[width=\unitlength,page=6]{images/3.pdf}}%
    \put(0.04022209,0.67595795){\makebox(0,0)[lt]{\lineheight{1.25}\smash{\begin{tabular}[t]{l}410\end{tabular}}}}%
    \put(0,0){\includegraphics[width=\unitlength,page=7]{images/3.pdf}}%
    \put(0.04022209,0.57905638){\makebox(0,0)[lt]{\lineheight{1.25}\smash{\begin{tabular}[t]{l}420\end{tabular}}}}%
    \put(0,0){\includegraphics[width=\unitlength,page=8]{images/3.pdf}}%
    \put(0.04022209,0.48215481){\makebox(0,0)[lt]{\lineheight{1.25}\smash{\begin{tabular}[t]{l}430\end{tabular}}}}%
    \put(0,0){\includegraphics[width=\unitlength,page=9]{images/3.pdf}}%
    \put(0.04022209,0.38525324){\makebox(0,0)[lt]{\lineheight{1.25}\smash{\begin{tabular}[t]{l}440\end{tabular}}}}%
    \put(0,0){\includegraphics[width=\unitlength,page=10]{images/3.pdf}}%
    \put(0.04022209,0.28835167){\makebox(0,0)[lt]{\lineheight{1.25}\smash{\begin{tabular}[t]{l}450\end{tabular}}}}%
    \put(0,0){\includegraphics[width=\unitlength,page=11]{images/3.pdf}}%
    \put(0.04022209,0.1914501){\makebox(0,0)[lt]{\lineheight{1.25}\smash{\begin{tabular}[t]{l}460\end{tabular}}}}%
    \put(0,0){\includegraphics[width=\unitlength,page=12]{images/3.pdf}}%
    \put(0.04022209,0.09454853){\makebox(0,0)[lt]{\lineheight{1.25}\smash{\begin{tabular}[t]{l}470\end{tabular}}}}%
    \put(0.0305288,0.35147222){\rotatebox{90}{\makebox(0,0)[lt]{\lineheight{1.25}\smash{\begin{tabular}[t]{l}Pixel\end{tabular}}}}}%
    \put(0,0){\includegraphics[width=\unitlength,page=13]{images/3.pdf}}%
    \put(0.78,0.36177807){\makebox(0,0)[lt]{\lineheight{1.25}\smash{\begin{tabular}[t]{l}Ground truth\end{tabular}}}}%
    \put(0,0){\includegraphics[width=\unitlength,page=14]{images/3.pdf}}%
    \put(0.78,0.32439742){\makebox(0,0)[lt]{\lineheight{1.25}\smash{\begin{tabular}[t]{l}Observation\end{tabular}}}}%
    \put(0,0){\includegraphics[width=\unitlength,page=15]{images/3.pdf}}%
    \put(0.78,0.28701678){\makebox(0,0)[lt]{\lineheight{1.25}\smash{\begin{tabular}[t]{l}CVM\end{tabular}}}}%
    \put(0,0){\includegraphics[width=\unitlength,page=16]{images/3.pdf}}%
    \put(0.78,0.24963613){\makebox(0,0)[lt]{\lineheight{1.25}\smash{\begin{tabular}[t]{l}Social-Impl.\end{tabular}}}}%
    \put(0,0){\includegraphics[width=\unitlength,page=17]{images/3.pdf}}%
    \put(0.78,0.21225549){\makebox(0,0)[lt]{\lineheight{1.25}\smash{\begin{tabular}[t]{l}AgentFormer\end{tabular}}}}%
    \put(0,0){\includegraphics[width=\unitlength,page=18]{images/3.pdf}}%
    \put(0.78,0.17487484){\makebox(0,0)[lt]{\lineheight{1.25}\smash{\begin{tabular}[t]{l}Ynet\end{tabular}}}}%
    \put(0,0){\includegraphics[width=\unitlength,page=19]{images/3.pdf}}%
    \put(0.78,0.1374942){\makebox(0,0)[lt]{\lineheight{1.25}\smash{\begin{tabular}[t]{l}Traj++\end{tabular}}}}%
    \put(0,0){\includegraphics[width=\unitlength,page=20]{images/3.pdf}}%
    \put(0.78,0.10011355){\makebox(0,0)[lt]{\lineheight{1.25}\smash{\begin{tabular}[t]{l}Sgan\end{tabular}}}}%
  \end{picture}%
\endgroup%

%% file: chapters/06_conclusion.tex
\section{Conclusion}
\label{sec_conc}
In this study, we have conducted a comprehensive evaluation of state-of-the-art pedestrian trajectory prediction methods when confronted with the requirements found in autonomous systems. Our analysis focused on measuring the ADE and FDE metrics based on single trajectories, investigating the impact of individual timesteps within the motion history, and measuring the overall runtime for different scenario sizes. Our findings demonstrate that Trajectron++ and Social-Implicit, which leverage graph-based interaction modeling, yield the most accurate results among the investigated architectures. Furthermore, we discovered that many models underutilize the available motion history with the majority of approaches only considering the first two timesteps.

When evaluating the trade-off between accuracy and runtime, Social-Implicit demonstrates the best performance overall, ranking second only to the CVM on the ETH/UCY dataset. This result can largely be attributed to the shortcomings of learning-based approaches to handle static scenarios. In addition, state-changes represent a challenge for all investigated approaches and require the inclusion of additional features such as semantic maps. Consequently, although the CVM provides a good approximation for most dynamic cases, improvements can still be achieved when considering a pedestrian's intention. Moving forward, this study offers valuable insights for three distinct groups: researchers in the autonomous driving domain intending to implement these algorithms into autonomous systems, scientists dedicated to enhancing existing approaches in terms of runtime and accuracy, and those seeking to devise a new metric that accurately represents the requisites for real-world applications. While future research directions are given by the development of hybrid approaches or the incorporation of intention recognition into the trajectory prediction framework, it remains an open question whether some pedestrian behavior might just not be predictable.

%% file: images/20230616_Fig_Runtime_over_Agents_YNet_v01.pdf_tex
\begingroup%
  \makeatletter%
  \providecommand\color[2][]{%
    \errmessage{(Inkscape) Color is used for the text in Inkscape, but the package 'color.sty' is not loaded}%
    \renewcommand\color[2][]{}%
  }%
  \providecommand\transparent[1]{%
    \errmessage{(Inkscape) Transparency is used (non-zero) for the text in Inkscape, but the package 'transparent.sty' is not loaded}%
    \renewcommand\transparent[1]{}%
  }%
  \providecommand\rotatebox[2]{#2}%
  \newcommand*\fsize{\dimexpr\f@size pt\relax}%
  \newcommand*\lineheight[1]{\fontsize{\fsize}{#1\fsize}\selectfont}%
  \ifx\svgwidth\undefined%
    \setlength{\unitlength}{234bp}%
    \ifx\svgscale\undefined%
      \relax%
    \else%
      \setlength{\unitlength}{\unitlength * \real{\svgscale}}%
    \fi%
  \else%
    \setlength{\unitlength}{\svgwidth}%
  \fi%
  \global\let\svgwidth\undefined%
  \global\let\svgscale\undefined%
  \makeatother%
  \begin{picture}(1,0.46210875)%
    \lineheight{1}%
    \setlength\tabcolsep{0pt}%
    \put(0,0){\includegraphics[width=\unitlength,page=1]{images/20230616_Fig_Runtime_over_Agents_YNet_v01.pdf}}%
    \put(0.12660256,0.06232253){\makebox(0,0)[lt]{\lineheight{1.25}\smash{\begin{tabular}[t]{l}0\end{tabular}}}}%
    \put(0.3707266,0.06232253){\makebox(0,0)[lt]{\lineheight{1.25}\smash{\begin{tabular}[t]{l}20\end{tabular}}}}%
    \put(0.62606827,0.06232253){\makebox(0,0)[lt]{\lineheight{1.25}\smash{\begin{tabular}[t]{l}40\end{tabular}}}}%
    \put(0.88141026,0.06232253){\makebox(0,0)[lt]{\lineheight{1.25}\smash{\begin{tabular}[t]{l}60\end{tabular}}}}%
    \put(0.35256442,0.00890362){\makebox(0,0)[lt]{\lineheight{1.25}\smash{\begin{tabular}[t]{l}Number of agents\end{tabular}}}}%
    \put(0,0){\includegraphics[width=\unitlength,page=2]{images/20230616_Fig_Runtime_over_Agents_YNet_v01.pdf}}%
    \put(0.10341891,0.0983267){\makebox(0,0)[lt]{\lineheight{1.25}\smash{\begin{tabular}[t]{l}0\end{tabular}}}}%
    \put(0.05854712,0.20302766){\makebox(0,0)[lt]{\lineheight{1.25}\smash{\begin{tabular}[t]{l}100\end{tabular}}}}%
    \put(0.05854712,0.3077283){\makebox(0,0)[lt]{\lineheight{1.25}\smash{\begin{tabular}[t]{l}200\end{tabular}}}}%
    \put(0.05854712,0.41242926){\makebox(0,0)[lt]{\lineheight{1.25}\smash{\begin{tabular}[t]{l}300\end{tabular}}}}%
    \put(0.04038462,0.13358311){\rotatebox{90}{\makebox(0,0)[lt]{\lineheight{1.25}\smash{\begin{tabular}[t]{l}Runtime in ms\end{tabular}}}}}%
    \put(0,0){\includegraphics[width=\unitlength,page=3]{images/20230616_Fig_Runtime_over_Agents_YNet_v01.pdf}}%
  \end{picture}%
\endgroup%

%% file: images/20230616_Fig_Runtime_over_Agents_Trajectron_v01.pdf_tex
\begingroup%
  \makeatletter%
  \providecommand\color[2][]{%
    \errmessage{(Inkscape) Color is used for the text in Inkscape, but the package 'color.sty' is not loaded}%
    \renewcommand\color[2][]{}%
  }%
  \providecommand\transparent[1]{%
    \errmessage{(Inkscape) Transparency is used (non-zero) for the text in Inkscape, but the package 'transparent.sty' is not loaded}%
    \renewcommand\transparent[1]{}%
  }%
  \providecommand\rotatebox[2]{#2}%
  \newcommand*\fsize{\dimexpr\f@size pt\relax}%
  \newcommand*\lineheight[1]{\fontsize{\fsize}{#1\fsize}\selectfont}%
  \ifx\svgwidth\undefined%
    \setlength{\unitlength}{234bp}%
    \ifx\svgscale\undefined%
      \relax%
    \else%
      \setlength{\unitlength}{\unitlength * \real{\svgscale}}%
    \fi%
  \else%
    \setlength{\unitlength}{\svgwidth}%
  \fi%
  \global\let\svgwidth\undefined%
  \global\let\svgscale\undefined%
  \makeatother%
  \begin{picture}(1,0.46210875)%
    \lineheight{1}%
    \setlength\tabcolsep{0pt}%
    \put(0,0){\includegraphics[width=\unitlength,page=1]{images/20230616_Fig_Runtime_over_Agents_Trajectron_v01.pdf}}%
    \put(0.14583333,0.06232253){\makebox(0,0)[lt]{\lineheight{1.25}\smash{\begin{tabular}[t]{l}0\end{tabular}}}}%
    \put(0.38354712,0.06232253){\makebox(0,0)[lt]{\lineheight{1.25}\smash{\begin{tabular}[t]{l}20\end{tabular}}}}%
    \put(0.63247853,0.06232253){\makebox(0,0)[lt]{\lineheight{1.25}\smash{\begin{tabular}[t]{l}40\end{tabular}}}}%
    \put(0.88141026,0.06232253){\makebox(0,0)[lt]{\lineheight{1.25}\smash{\begin{tabular}[t]{l}60\end{tabular}}}}%
    \put(0.36217981,0.00890362){\makebox(0,0)[lt]{\lineheight{1.25}\smash{\begin{tabular}[t]{l}Number of agents\end{tabular}}}}%
    \put(0,0){\includegraphics[width=\unitlength,page=2]{images/20230616_Fig_Runtime_over_Agents_Trajectron_v01.pdf}}%
    \put(0.12264968,0.0983267){\makebox(0,0)[lt]{\lineheight{1.25}\smash{\begin{tabular}[t]{l}0\end{tabular}}}}%
    \put(0.07777788,0.25537798){\makebox(0,0)[lt]{\lineheight{1.25}\smash{\begin{tabular}[t]{l}500\end{tabular}}}}%
    \put(0.05854712,0.41242926){\makebox(0,0)[lt]{\lineheight{1.25}\smash{\begin{tabular}[t]{l}1000\end{tabular}}}}%
    \put(0.04038462,0.13358311){\rotatebox{90}{\makebox(0,0)[lt]{\lineheight{1.25}\smash{\begin{tabular}[t]{l}Runtime in ms\end{tabular}}}}}%
    \put(0,0){\includegraphics[width=\unitlength,page=3]{images/20230616_Fig_Runtime_over_Agents_Trajectron_v01.pdf}}%
  \end{picture}%
\endgroup%

%% file: main.bbl
\begin{thebibliography}{10}
\providecommand{\url}[1]{#1}
\csname url@samestyle\endcsname
\providecommand{\newblock}{\relax}
\providecommand{\bibinfo}[2]{#2}
\providecommand{\BIBentrySTDinterwordspacing}{\spaceskip=0pt\relax}
\providecommand{\BIBentryALTinterwordstretchfactor}{4}
\providecommand{\BIBentryALTinterwordspacing}{\spaceskip=\fontdimen2\font plus
\BIBentryALTinterwordstretchfactor\fontdimen3\font minus
  \fontdimen4\font\relax}
\providecommand{\BIBforeignlanguage}[2]{{%
\expandafter\ifx\csname l@#1\endcsname\relax
\typeout{** WARNING: IEEEtran.bst: No hyphenation pattern has been}%
\typeout{** loaded for the language `#1'. Using the pattern for}%
\typeout{** the default language instead.}%
\else
\language=\csname l@#1\endcsname
\fi
#2}}
\providecommand{\BIBdecl}{\relax}
\BIBdecl

\bibitem{rasouli_autonomous_2018}
A.~Rasouli and J.~K. Tsotsos, ``Autonomous vehicles that interact with
  pedestrians: A survey of theory and practice.''

\bibitem{gorrini_observation_2018}
A.~Gorrini, L.~Crociani, G.~Vizzari, and S.~Bandini, ``Observation results on
  pedestrian-vehicle interactions at non-signalized intersections towards
  simulation,'' vol.~59, pp. 269--285.

\bibitem{who_report}
\BIBentryALTinterwordspacing
{World Health Organization (WHO)}, ``Global status report on road safety 2018:
  Summary,'' December 2018, [last accessed 09-Oct-2023]. [Online]. Available:
  \url{https://www.who.int/publications/i/item/WHO-NMH-NVI-18.20}
\BIBentrySTDinterwordspacing

\bibitem{mohamed_social-implicit_2022}
A.~Mohamed, D.~Zhu, W.~Vu, M.~Elhoseiny, and C.~Claudel, ``Social-implicit:
  Rethinking trajectory prediction evaluation and the effectiveness of implicit
  maximum likelihood estimation.''

\bibitem{yue_human_2022}
J.~Yue, D.~Manocha, and H.~Wang, ``Human trajectory prediction via neural
  social physics.''

\bibitem{zhang_forceformer_2023}
W.~Zhang, H.~Cheng, F.~T. Johora, and M.~Sester, ``{ForceFormer}: Exploring
  social force and transformer for pedestrian trajectory prediction.''

\bibitem{fang_behavioral_2022}
J.~Fang, F.~Wang, P.~Shen, Z.~Zheng, J.~Xue, and T.-s. Chua, ``Behavioral
  intention prediction in driving scenes: A survey.''

\bibitem{sharma_pedestrian_2022}
N.~Sharma, C.~Dhiman, and S.~Indu, ``Pedestrian intention prediction for
  autonomous vehicles: A comprehensive survey,'' vol. 508, pp. 120--152.

\bibitem{rudenko_human_2020}
\BIBentryALTinterwordspacing
A.~Rudenko, L.~Palmieri, M.~Herman, K.~M. Kitani, D.~M. Gavrila, and K.~O.
  Arras, ``Human motion trajectory prediction: a survey,'' vol.~39, no.~8, pp.
  895--935, publisher: {SAGE} Publications Ltd {STM}. [Online]. Available:
  \url{https://doi.org/10.1177/0278364920917446}
\BIBentrySTDinterwordspacing

\bibitem{helbing_social_1995}
D.~Helbing and P.~Molnar, ``Social force model for pedestrian dynamics,''
  vol.~51, no.~5, pp. 4282--4286.

\bibitem{gupta_social_2018}
A.~Gupta, J.~Johnson, L.~Fei-Fei, S.~Savarese, and A.~Alahi, ``Social {GAN}:
  Socially acceptable trajectories with generative adversarial networks.''

\bibitem{salzmann_trajectron_2021}
T.~Salzmann, B.~Ivanovic, P.~Chakravarty, and M.~Pavone, ``Trajectron++:
  Dynamically-feasible trajectory forecasting with heterogeneous data.''

\bibitem{lerner_crowds_2007}
A.~Lerner, Y.~Chrysanthou, and D.~Lischinski, ``Crowds by example,'' vol.~26,
  no.~3, pp. 655--664, \_eprint:
  https://onlinelibrary.wiley.com/doi/pdf/10.1111/j.1467-8659.2007.01089.x.

\bibitem{pellegrini_youll_2009}
S.~Pellegrini, A.~Ess, K.~Schindler, and L.~van Gool, ``You'll never walk
  alone: Modeling social behavior for multi-target tracking,'' in \emph{2009
  {IEEE} 12th International Conference on Computer Vision}, pp. 261--268,
  {ISSN}: 2380-7504.

\bibitem{charalambous_learning_2016}
P.~Charalambous and Y.~Chrysanthou, ``Learning heterogeneous crowd behavior
  from the real world.''

\bibitem{caesar_nuscenes_2020}
H.~Caesar, V.~Bankiti, A.~H. Lang, S.~Vora, V.~E. Liong, Q.~Xu, A.~Krishnan,
  Y.~Pan, G.~Baldan, and O.~Beijbom, ``{nuScenes}: A multimodal dataset for
  autonomous driving.''

\bibitem{scholler_what_2020}
C.~Schöller, V.~Aravantinos, F.~Lay, and A.~Knoll, ``What the constant
  velocity model can teach us about pedestrian motion prediction.''

\bibitem{sun_human_nodate}
J.~Sun, Y.~Li, L.~Chai, H.-S. Fang, Y.-L. Li, and C.~Lu, ``Human trajectory
  prediction with momentary observation.''

\bibitem{van_den_berg_reciprocal_2011}
J.~van~den Berg, S.~J. Guy, M.~Lin, and D.~Manocha, ``Reciprocal n-body
  collision avoidance,'' in \emph{Robotics Research}, ser. Springer Tracts in
  Advanced Robotics, C.~Pradalier, R.~Siegwart, and G.~Hirzinger, Eds.\hskip
  1em plus 0.5em minus 0.4em\relax Springer, pp. 3--19.

\bibitem{kim_brvo_2015}
\BIBentryALTinterwordspacing
S.~Kim, S.~J. Guy, W.~Liu, D.~Wilkie, R.~W. Lau, M.~C. Lin, and D.~Manocha,
  ``{BRVO}: Predicting pedestrian trajectories using velocity-space
  reasoning,'' vol.~34, no.~2, pp. 201--217, publisher: {SAGE} Publications Ltd
  {STM}. [Online]. Available: \url{https://doi.org/10.1177/0278364914555543}
\BIBentrySTDinterwordspacing

\bibitem{alahi_social_2016}
A.~Alahi, K.~Goel, V.~Ramanathan, A.~Robicquet, L.~Fei-Fei, and S.~Savarese,
  ``Social {LSTM}: Human trajectory prediction in crowded spaces,'' in
  \emph{2016 {IEEE} Conference on Computer Vision and Pattern Recognition
  ({CVPR})}, pp. 961--971, {ISSN}: 1063-6919.

\bibitem{mangalam_goals_2020}
K.~Mangalam, Y.~An, H.~Girase, and J.~Malik, ``From goals, waypoints \& paths
  to long term human trajectory forecasting.''

\bibitem{yuan_agentformer_2021}
Y.~Yuan, X.~Weng, Y.~Ou, and K.~Kitani, ``{AgentFormer}: Agent-aware
  transformers for socio-temporal multi-agent forecasting.''

\bibitem{sighencea_review_2021}
B.~Sighencea, R.~Stanciu, and C.~Căleanu, ``A review of deep learning-based
  methods for pedestrian trajectory prediction,'' vol.~21, p. 7543.

\bibitem{NEURIPS2019_9015}
A.~Paszke, S.~Gross, F.~Massa, A.~Lerer, J.~Bradbury, G.~Chanan, T.~Killeen,
  Z.~Lin, N.~Gimelshein, L.~Antiga, A.~Desmaison, A.~Kopf, E.~Yang, Z.~DeVito,
  M.~Raison, A.~Tejani, S.~Chilamkurthy, B.~Steiner, L.~Fang, J.~Bai, and
  S.~Chintala, ``Pytorch: An imperative style, high-performance deep learning
  library,'' in \emph{Advances in Neural Information Processing Systems
  32}.\hskip 1em plus 0.5em minus 0.4em\relax Curran Associates, Inc., 2019,
  pp. 8024--8035.

\bibitem{kothari_human_2022}
P.~Kothari, S.~Kreiss, and A.~Alahi, ``Human trajectory forecasting in crowds:
  A deep learning perspective,'' vol.~23, no.~7, pp. 7386--7400, conference
  Name: {IEEE} Transactions on Intelligent Transportation Systems.

\bibitem{kitti2012}
A.~Geiger, P.~Lenz, and R.~Urtasun, ``Are we ready for autonomous driving? the
  kitti vision benchmark suite,'' in \emph{2012 IEEE Conference on Computer
  Vision and Pattern Recognition}, 2012, pp. 3354--3361.

\bibitem{sohn_learning_2015}
K.~Sohn, H.~Lee, and X.~Yan, ``Learning structured output representation using
  deep conditional generative models,'' in \emph{Advances in Neural Information
  Processing Systems}, C.~Cortes, N.~Lawrence, D.~Lee, M.~Sugiyama, and
  R.~Garnett, Eds., vol.~28.\hskip 1em plus 0.5em minus 0.4em\relax Curran
  Associates, Inc.

\bibitem{vizzari_agent_2020}
G.~Vizzari, L.~Crociani, and S.~Bandini, ``An agent-based model for plausible
  wayfinding in pedestrian simulation,'' \emph{Engineering Applications of
  Artificial Intelligence}, vol.~87, p. 103241, 2020.

\bibitem{schadschneider_evacuation_2009}
\BIBentryALTinterwordspacing
A.~Schadschneider, W.~Klingsch, H.~Kl{\"u}pfel, T.~Kretz, C.~Rogsch, and
  A.~Seyfried, \emph{Evacuation Dynamics: Empirical Results, Modeling and
  Applications}.\hskip 1em plus 0.5em minus 0.4em\relax New York, NY: Springer
  New York, 2009, pp. 3142--3176. [Online]. Available:
  \url{https://doi.org/10.1007/978-0-387-30440-3_187}
\BIBentrySTDinterwordspacing

\bibitem{rasouli_pie_2019}
A.~Rasouli, I.~Kotseruba, T.~Kunic, and J.~Tsotsos, ``{PIE}: A large-scale
  dataset and models for pedestrian intention estimation and trajectory
  prediction,'' in \emph{2019 {IEEE}/{CVF} International Conference on Computer
  Vision ({ICCV})}.\hskip 1em plus 0.5em minus 0.4em\relax {IEEE}, pp.
  6261--6270.

\bibitem{robicquet_learning_2016}
A.~Robicquet, A.~Sadeghian, A.~Alahi, and S.~Savarese, ``Learning social
  etiquette: Human trajectory understanding in crowded scenes,'' in
  \emph{Computer Vision – {ECCV} 2016}, ser. Lecture Notes in Computer
  Science, B.~Leibe, J.~Matas, N.~Sebe, and M.~Welling, Eds.\hskip 1em plus
  0.5em minus 0.4em\relax Springer International Publishing, pp. 549--565.

\bibitem{ivanovic_expanding_2023}
B.~Ivanovic, J.~Harrison, and M.~Pavone, ``Expanding the deployment envelope of
  behavior prediction via adaptive meta-learning.''

\end{thebibliography}
